\documentclass{article}
\usepackage{amsmath, amssymb, amsthm, geometry, graphicx, listings, verbatim}
\usepackage{fancyvrb}
\usepackage[utf8]{inputenc}
\usepackage{pmboxdraw}
\usepackage{subcaption}
\usepackage{graphicx}
\usepackage[utf8]{inputenc}
\usepackage{listings} 
\usepackage{xcolor}   

\usepackage{booktabs} 
\usepackage{graphicx}
\usepackage[
    pdftitle={Federated Learning and Class Imbalances},
    pdfauthor={Siqi Zhu},
    pdfsubject={Federated Learning, Machine Learning},
    pdfkeywords={federated learning, label noise, heterogeneous models}
]{hyperref}
\usepackage{pgfplots}
\usepackage{graphicx} 
\usepackage{algorithm}
\usepackage{algorithmic}
\usepackage{booktabs}
\usepackage{geometry}
\usepackage{longtable}
\usepackage{array}
\usepackage{tikz}
\begin{document}

\title{Federated Learning and Class Imbalances: \\
Robust Heterogeneous Federated Learning with Label Noise}

\author{Siqi Zhu\thanks{Department of Applied Mathematics and Theoretical Physics, University of Cambridge. Email: \texttt{sz478@cam.ac.uk}} \and Joshua D. Kaggie\thanks{Department of Radiology, University of Cambridge. Email: \texttt{jk636@medschl.cam.ac.uk}}}

\date{November 2025}
\maketitle
\thispagestyle{empty}

\begin{abstract}
Federated Learning (FL) enables collaborative model training across decentralized devices while preserving data privacy. However, real-world FL deployments face critical challenges such as data imbalances, including label noise and non-IID distributions. RHFL+, a state-of-the-art method, was proposed to address these challenges in settings with heterogeneous client models. This work investigates the robustness of RHFL+ under class imbalances through three key contributions: (1) reproduction of RHFL+ along with all benchmark algorithms under a unified evaluation framework; (2) extension of RHFL+ to real-world medical imaging datasets, including CBIS-DDSM~\cite{lee2016cbisddsm,lee_curated_2017}, BreastMNIST~\cite{yang2023medmnist} and BHI~\cite{mooney2017breast}; (3) a novel implementation using NVFlare~\cite{nvflare}, NVIDIA's production-level federated learning framework, enabling a modular, scalable and deployment-ready codebase. To validate effectiveness, extensive ablation studies, algorithmic comparisons under various noise conditions and scalability experiments across increasing numbers of clients are conducted.
\end{abstract}

\section{Introduction}
\label{}

\subsection{Why Federated Learning?}

\hspace{1.5em}Federated Learning is a collaborative machine learning paradigm that allows multiple devices to train collaboratively without centralizing their raw data~\cite{ludwig2022federated}. Unlike traditional machine learning where all data is uploaded to a server, FL enables clients to keep data local and only exchange model parameters or updates. 

It was initially introduced by Google \cite{konecny2016federatedoptimization, DBLP:journals/corr/KonecnyMYRSB16} to address concerns about data privacy in mobile environments. The well-known Federated Averaging (FedAvg) algorithm was later proposed in \cite{pmlr-v54-mcmahan17a}, allowing efficient on-device training for applications such as Google Assistant.

FL has then gained attention due to its ability to preserve privacy, reduce communication with central servers~\cite{gabrielli2023survey} and leverage data stored across distributed devices, such as smartphones, IoT sensors and medical imaging systems. By offloading computation to edge devices, FL reduces the burden on centralized infrastructure and enables learning from sensitive data, which is particularly important in healthcare and other privacy-sensitive domains.

Originally, FL relied on a centralized architecture in which a global server coordinated training by aggregating model updates from clients and broadcasting the global model~\cite{pmlr-v54-mcmahan17a}. However, this setup assumes homogeneous model architectures across clients. To address these limitations, heterogeneous FL approaches have emerged, allowing clients to train models with different architectures tailored to their computational capabilities and data characteristics without relying on a central model. This flexibility is crucial in real-world scenarios, such as IoT networks and mobile devices, where system heterogeneity is the norm~\cite{yi2023fedghheterogeneousfederatedlearning}. Techniques such as knowledge distillation and collaborative representation learning have been proposed to bridge the differences between heterogeneous models~\cite{guo2024comprehensive}.

FL has seen successful applications across a wide range of domains. One early deployment was Google's Gboard, where FL was used to improve keyboard predictions without uploading users' text data~\cite{hard2019federatedlearningmobilekeyboard}. In healthcare, FL enables collaborative training across hospitals without sharing patient data, facilitating improved models for disease diagnosis~\cite{xu2021federated}. In industrial systems, FL supports predictive maintenance and anomaly detection for security monitoring by learning from distributed sensor data~\cite{jiang2024blockchained}.

These advances demonstrate that FL is evolving beyond its initial centralized design toward more decentralized, heterogeneous and real-world-compatible learning paradigms. Several comprehensive surveys have reviewed the evolution, applications and limitations of FL \cite{banabilah_federated_2022, wen_survey_2023, guo2024comprehensive}. 

\subsection{Challenges}

\hspace{1.5em}Despite the potential of FL, several key challenges remain. One major issue is the communication overhead. Frequent exchange of model updates between clients and the server, especially in high-dimensional models, incurs significant bandwidth consumption and latency. This is problematic, especially on resource-limited devices~\cite{li2020federated, kairouz2021advancesopenproblemsfederated}. Another challenge is system heterogeneity, where clients differ in computational power, memory capacity, network connectivity and availability. Such variations result in imbalanced training progress and pose difficulties in coordinating synchronous updates. Additionally, statistical heterogeneity arises because data across clients is often non-IID, making model convergence slower and potentially biased towards dominant clients~\cite{zhu2021federated}. A further challenge is the presence of label noise in local datasets. In practice, labels may be corrupted due to human error or inherent biases in the data collection process. This problem is worse in decentralized settings where the server lacks direct access to client data and thus cannot identify mislabels~\cite{10816157}. Noisy labels not only degrade model quality but are also particularly difficult to manage in scenarios with model heterogeneity, where different clients use different architectures. These limitations motivate the development of robust and flexible FL algorithms such as RHFL+~\cite{10816157}.

\subsection{Class Imbalances}

\hspace{1.5em}In the context of FL, class imbalances refer to deviations from ideal, clean and balanced class distributions in the clients' local datasets. These imbalances can arise from a variety of factors, most notably the non-IID nature of data across clients. One typical form is when the number of samples per class varies significantly either within a client or across the federation. Another critical, but often under-addressed form of imbalances is label noise, which refers to incorrectly assigned class labels in the training data.

While a dataset may appear balanced in terms of raw class counts, mislabeled samples distort the true distribution of the data, implicitly. In FL, this becomes more problematic due to the decentralized nature of training: each client trains locally, often with limited data and the central server cannot inspect raw samples due to privacy constraints. Consequently, mislabeled data can corrupt the local model and propagate wrong updates to the global model silently \cite{10816157, wang2022fednoil, xu2022fedcorr}.

In this thesis, label noise is the primary form of class imbalances under investigation. By manipulating both the \textit{noise type} (e.g. symmetric or asymmetric noise) and the \textit{noise rate} (i.e., the proportion of corrupted labels), we simulate realistic conditions where clients may experience varying levels of annotation errors. This approach provides insights into the resilience of FL algorithms against noisy annotations, particularly in non-IID scenarios and highlights the necessity of explicitly accounting for label quality in FL research.

\subsection{State-of-the-Art Benchmarks}

\hspace{1.5em}This thesis follows the experimental setup of the RHFL+ paper~\cite{10816157} and includes all methods listed in Table~\ref{tab:fl_comparison} as benchmarks. RHFL+ is a recent state-of-the-art method in federated learning that simultaneously addresses model heterogeneity and label noise. It is explained in detail later in the next Section. RHFL+ , as stated in its paper, outperforms previous methods such as FedMD~\cite{li2019fedmdheterogenousfederatedlearning}, FedDF~\cite{lin2020ensemble} and KT-pFL~\cite{zhang2021parameterizedknowledgetransferpersonalized} under varying levels of label corruption. For example, under 20\% symmetric label noise, RHFL+ achieves higher stability and accuracy compared to all baselines across both homogeneous and heterogeneous client settings~\cite{10816157}. This robustness makes RHFL+ an ideal primary subject of evaluation in this thesis.

\subsection{Goal}

\hspace{1.5em}While many papers propose and implement novel FL algorithms, their experimental setups often rely on centralized simulation or simplified local emulation, failing to reflect realistic deployment conditions. This thesis aims to reproduce and validate the results of RHFL+~\cite{10816157} in a truly distributed environment using NVFlare~\cite{nvflare}, NVIDIA's open source federated learning platform. 

The primary objective is to faithfully reproduce the performance evaluations of RHFL+ while ensuring that the implementation is compatible with real-world federated environments. In addition, this work seeks to develop a modular, transferable and extensible NVFlare-based framework to support further research across diverse FL settings. A further goal is to apply and adapt RHFL+ to a medical imaging use case, evaluating its effectiveness under data scarcity and label imbalance, as commonly observed in healthcare applications.

\subsection{Structure of Thesis}

\hspace{1.5em}The remainder of this thesis is organized as follows: Section~\ref{sec:overview} provides background knowledge on federated learning and the challenges posed by label noise. Section~\ref{sec:reproduce} details the reproduction of RHFL+ using NVFlare, including the experimental setup, design challenges and an analysis of the reproduced results. Section~\ref{sec:extension} explores the extension of other CIFAR-10 experiments, as well as applying RHFL+ to medical imaging datasets, evaluating its performance in the context of domain-specific constraints such as data scarcity and label imbalance. Section~\ref{sec:conclusion} summarizes the key findings derived from all experiments and their implications. Finally, Section~\ref{sec:next} discusses the limitations of the current approach and outlines potential directions for future work.

\section{Overview of Federated Learning}
\label{sec:overview}

\subsection{Problem Formulation}

\hspace{1.5em}
Let the total number of participating clients be denoted by $n$, where clients are indexed as $i = 1, 2, \dots, n$. Each client $i$ holds a private dataset $\mathcal{D}_i$, where $\tilde{y}_i$ is a potentially corrupted version of the true label $y_i$. The label noise at each client is characterized by a noise rate $\eta_k \in [0, 1]$, which defines the probability that $\tilde{y}_i \neq y_i$. Each client $i$ also maintains its own model parameters $\boldsymbol{\theta}_i$. Depending on the FL setting, $\boldsymbol{\theta}_i$ may represent the same model architecture across all clients (homogeneous FL). Or, different architectures (heterogeneous FL), in which case only the output representations (e.g. logits) might need to be aligned.

The major differences between FL and other machine learning paradigms are summarized in Table~\ref{tab:fl_and_ml}.

\begin{table}[h]
\centering
\begin{tabular}{|l|c|c|c|c|}
\hline
\textbf{Type} & \textbf{Central Server} & \textbf{Data Sharing} & \textbf{Data Distribution} \\
\hline
Centralized ML       & Yes & Raw data to server & Any               \\
Distributed ML       & Yes & Shared dataset     & Usually IID      \\
Decentralized ML     & No  & No central data    & Usually IID        \\
Federated Learning   & Yes & No data sharing & Often non-IID    \\
\hline
\end{tabular}
\caption{Comparison of FL with other ML training paradigms  
\cite{banabilah_federated_2022}}
\label{tab:fl_and_ml}
\end{table}

\subsection{Categorization}

\paragraph{Homogeneous Federated Learning}

Early FL algorithms operate under the assumption that all clients share an identical model architecture, which ensures that model parameters are aligned for straightforward, parameter-wise aggregation on the server\cite{makhija2022architectureagnosticfederatedlearning}. Examples of such algorithms include the classical FedAvg \cite{pmlr-v54-mcmahan17a}, Each client trains locally on its private dataset and periodically sends updates to the server, which aggregates them (by averaging in this case) into a new global model. The detailed steps are shown in Algorithm \ref{alg:fedavg}.

\begin{algorithm}[H]
\caption{Federated Averaging (FedAvg)}
\label{alg:fedavg}
\begin{algorithmic}[1]
\REQUIRE Number of clients $n$, local datasets $\{\mathcal{D}_i\}_{i=1}^{n}$, number of rounds $T$
\STATE Initialize global model parameters $\boldsymbol{\theta}^{(0)}$
\FOR{each round $t = 0, 1, \dots, T-1$}
    \STATE Server selects a subset of clients $\mathcal{S}^{(t)} \subseteq \{1, 2, \dots, n\}$
    \FOR{each client $i \in \mathcal{S}^{(t)}$ \textbf{in parallel}}
        \STATE Client $i$ receives $\boldsymbol{\theta}^{(t)}$
        \STATE Client $i$ updates model locally to obtain $\boldsymbol{\theta}_i^{(t+1)}$ using $\mathcal{D}_i$
    \ENDFOR
    \STATE Server aggregates the updates:
    \[
    \boldsymbol{\theta}^{(t+1)} = \sum_{i \in \mathcal{S}^{(t)}} \frac{|\mathcal{D}_i|}{\sum_{j \in \mathcal{S}^{(t)}} |\mathcal{D}_j|} \boldsymbol{\theta}_i^{(t+1)}
    \]
\ENDFOR
\end{algorithmic}
\label{algo:fedavg}
\end{algorithm}

Numerous extensions of FedAvg have been proposed to improve optimization under data and system heterogeneity. These variants introduce changes to the local training objective, aggregation rule, or update procedure while still assuming a homogeneous model across clients. For example, FedProx \cite{li2020federatedoptimizationheterogeneousnetworks} adds a proximal term to stabilize training under data heterogeneity. 

\subsubsection*{Heterogeneous Federated Learning}

\hspace{1.5em}In real-world scenarios, clients may operate in very different environments and use entirely different model architectures, due to differences in data types, computational resources, or application domains. This motivates heterogeneous FL, which relaxes the requirement of identical model structures by exchanging knowledge in forms other than raw model parameters. For example, FedMD\cite{li2019fedmdheterogenousfederatedlearning} allows clients to have different model architectures by having them share and distill soft logits (predicted probabilities) on a public auxiliary dataset, instead of sharing weights. Details can be found in Algorithm \ref{algo:fedmd}. Such distillation-based frameworks enable federated learning without requiring aligned model parameters, addressing a key limitation of homogeneous FL algorithms.

The key limitation of this method is that the output spaces (e.g. logits) have to be compatible, so that collaborative knowledge transfer is still possible. Aggregation is no longer done in parameter space, but instead in output space, using techniques such as logit distillation\cite{hinton2015distilling}, prototype matching\cite{snell2017prototypical}, or feature representation alignment\cite{romero2015fitnetshintsdeepnets}.

\begin{algorithm}[H]
\caption{General Framework for Heterogeneous FL}
\label{alg:hetero_fl}
\begin{algorithmic}[1]
\REQUIRE Number of clients $n$, local datasets $\{\mathcal{D}_i\}_{i=1}^{n}$, public dataset $\mathcal{D}_{\text{pub}}$, number of rounds $T$
\FOR{each round $t = 0, 1, \dots, T-1$}
    \FOR{each client $i$ \textbf{in parallel}}
        \STATE Train local model on private data $\mathcal{D}_i$
        \STATE Compute outputs (e.g. logits) on public data
        \STATE Send outputs to server
    \ENDFOR
    \STATE Server aggregates collaborative knowledge (e.g. by averaging logits or computing confidence weights)
    \FOR{each client $i$}
        \STATE Refine local model using collaborative loss (e.g. KL divergence)
    \ENDFOR
\ENDFOR
\end{algorithmic}
\label{algo:fedmd}
\end{algorithm}

Table~\ref{tab:fl_comparison} summarizes key differences between benchmark FL methods this thesis covers in terms of heterogeneity, aggregation, central model use, reliance on public data, share of prototypes or logits and distillation mechanism.

\begin{table}[htbp]
\centering
\scriptsize
\renewcommand{\arraystretch}{2}
\resizebox{\textwidth}{!}{%
\begin{tabular}{|l|c|c|c|c|c|c|}
\hline
\textbf{Method} & \textbf{Heterogeneous} & \textbf{Aggregation} & \textbf{Central Model} & \textbf{Public Data} & \textbf{Prototype/Logit} & \textbf{Distillation} \\
\hline
FedAvg\cite{pmlr-v54-mcmahan17a}     & No  & Weight-based                 & Yes                  & No  & None                   & No  \\
FedMD\cite{li2019fedmdheterogenousfederatedlearning} & Yes & Output-based & No & Yes & Logits & Yes\\
FedDF\cite{lin2020ensemble} & Yes & Output-based                 & Yes (per-client)     & Yes & Logits                 & Yes \\
KT-pFL\cite{zhang2021parameterizedknowledgetransferpersonalized}     & Yes & Output-based + Personalized  & No                   & Yes & Logits + C Matrix      & Yes \\
FedProto\cite{tan2022fedprotofederatedprototypelearning}   & Yes & Prototype-based              & Yes (Prototypes)     & No & Feature Prototypes     & No  \\
FCCL\cite{9879190}       & Yes & Output-based                 & No                   & Yes & Logits                 & Yes \\
FedGH\cite{yi2023fedghheterogeneousfederatedlearning}      & Yes & Representation-based         & Yes (Global Header)  & No  & Local Averaged Reps    & Yes \\
FedTGP\cite{zhang2024fedtgptrainableglobalprototypes}    & Yes & Prototype + Contrastive      & Yes (Trainable Prot.)& No & Prototypes             & Contrastive \\
RHFL\cite{fang2022robust}       & Yes & Output-based (SL loss)       & No                   & Yes & Soft Labels            & Yes \\
RHFL+\cite{10816157}      & Yes & Output + Confidence Weighting& No                   & Yes & Soft Labels + Confidence & Yes \\
AugHFL\cite{10378155}     & Yes & Output (Reliability Weighted)& No                   & Yes & Logits                 & Yes \\
\hline
\end{tabular}
}
\caption{Comparison of Federated Learning algorithms}
\label{tab:fl_comparison}
\end{table}

\section{Reproduction}
\label{sec:reproduce}

\hspace{1.5em}The original RHFL+ paper \cite{10816157} introduces a robust federated learning framework designed for model-heterogeneous scenarios under label noise. While the goal of this work is not to improve RHFL+ itself, I aim to rigorously validate its effectiveness and ensure reproducibility of its core results. To this end, I have reproduced all key experiments from the original paper, including ablation studies and comparative evaluations across various algorithms and noise settings.

It is important to highlight that the official codebase only provides an implementation for RHFL+ and does not include the baseline methods listed in the evaluation tables. As such, I developed a fully modular and extensible codebase that implements all benchmark methods from scratch, based on formulations and hyperparameters described in their original publications.

\subsection{Dynamic Local Noise Learning}

\hspace{1.5em}To mitigate the impact of label noise during client-side training, RHFL+ integrates Dynamic Label Refinement (DLR) with Symmetric Cross-Entropy (SL) loss. 

\paragraph{DLR\cite{10816157}}
generates adaptive soft labels by combining the noisy ground-truth label $\tilde{y}_i^k$ with the model prediction $\phi(x_i^k)$:

\begin{equation}
\hat{y}_i^k = (1 - s_{t_c}) \cdot \tilde{y}_i^k + s_{t_c} \cdot \phi(x_i^k),
\end{equation}

\noindent where the weight $s_{t_c}$ increases over training epochs $t_c$:

\begin{equation}
s_{t_c} = \frac{t_c}{\zeta T_c + t_c},
\end{equation}

\noindent with $\zeta$ being a hyperparameter that controls the influence of the prediction. 

\paragraph{SL\cite{wang2019symmetriccrossentropyrobust}}
is defined as:

\begin{equation}
\mathcal{L}_{\text{SL}} = \lambda \cdot \mathcal{L}_{\text{CE}} + \gamma \cdot \mathcal{L}_{\text{RCE}},
\end{equation}

\noindent where:

\begin{align}
\mathcal{L}_{\text{CE}} &= -\sum_{k=1}^{C} p(k|x) \log q(k|x), \\
\mathcal{L}_{\text{RCE}} &= -\sum_{k=1}^{C} q(k|x) \log p(k|x),
\end{align}

\noindent with $p(k|x)$ denoting the noisy label distribution and $q(k|x)$ the predicted distribution. This formulation helps prevent overfitting on the simple classes, while ensuring that hard-to-learn classes are still sufficiently trained.

\paragraph{Local Update}
Finally, the local model update step under DLR and SL loss becomes:

\begin{equation}
\theta_k^{t_l} \leftarrow \theta_k^{t_l-1} - \alpha \nabla_\theta \mathcal{L}_{sl}^{k, t_l-1}(\phi_\tau(x^k), \hat{y}^k),
\end{equation}

\noindent where $\alpha$ is the learning rate. This dynamic learning mechanism allows RHFL+ to effectively suppress noisy labels and improve model generalization across heterogeneous federated clients.

\subsection{Enhanced Client Confidence Reweighting (ECCR)}

\hspace{1.5em}This method dynamically adjusts the contribution of each client during the collaborative learning phase by jointly assessing two key factors: label quality and learning efficiency.

\paragraph{Label Quality}
The label quality of a client's local dataset $\widetilde{D}_k$ at communication round $t_c$ is estimated as:
\begin{equation}
Q^{t_c}(\widetilde{D}_k) = \left( \frac{1}{N_k} \sum_{i=1}^{N_k} L_{\text{sl}}^{k,t_c}(f(x_i^k, \theta_k), \widetilde{y}_i^k) \right)^{-1},
\end{equation}
where $L_{\text{sl}}$ reflects the agreement between predicted and noisy labels. Lower SL loss implies better label quality.

\paragraph{Learning Efficiency}
The evaluation of learning efficiency considers both the SL loss drop and the normalized model updates:
\begin{equation}
\mathcal{P}(\theta_k^{t_c}) = \frac{\Delta \mathcal{L}_{sl}^{k,t_c}}{\left( \frac{\Delta \theta_k^{t_c}}{|\theta_k|} + 1 \right)},
\label{eq:le}
\end{equation}
where $\Delta L_{\text{sl}}^{k,t_c}$ measures the improvement in SL loss and the denominator captures the normalized model updates to penalize instability or overfitting.

\paragraph{Client Confidence}
\begin{equation}
F_k^{t_c} = \widetilde{Q}^{t_c}(\widetilde{D}_k) \cdot P(\theta_k^{t_c}),
\end{equation}
which reflects both the trustworthiness of a client's data and its learning progress. 

\paragraph{Client Weight}
\begin{equation}
w_{e}^{k, t_c} = \frac{1}{K - 1} + \eta \cdot \frac{\mathcal{F}_{k}^{t_c}}{\sum_{k=1}^{K} |\mathcal{F}_{k}^{t_c}|},
\end{equation}
where $\eta$ controls the sensitivity to confidence scores. These raw weights are then normalized across all clients:
\begin{equation}
\mathcal{W}_{k}^{t_c} = \frac{w_{e}^{k, t_c}}{\sum_{k=1}^{K} w_{e}^{k, t_c}}.
\end{equation}

Finally, the calculation of client-specific collaborative loss is re-weighted using these confidence-adjusted weights. This reweighting strategy allows cleaner and more effective clients to play a greater role in global knowledge aggregation, thereby enhancing robustness and convergence of the federated learning system.

\begin{algorithm}[H]
\caption{RHFL+\cite{10816157}}
\label{alg:rhfl_plus}
\begin{algorithmic}[1]
\REQUIRE Number of clients $n$, local datasets $\{\mathcal{D}_i\}_{i=1}^{n}$, public dataset $\mathcal{D}_{\text{pub}}$, local models $\{\theta_i\}$, number of rounds $T$
\FOR{each round $t = 0, \dots, T-1$}
    \FOR{each client $i$ \textbf{in parallel}}
        \STATE \textbf{Phase 1 (Inference only)}:
        \STATE \quad Evaluate $\theta_i^{(t)}$ on $\mathcal{D}_i$ to estimate label quality $Q_i^{(t)}$ and learning efficiency $P_i^{(t)}$
        \STATE \quad Compute client confidence $F_{i}^{(t)}$ and upload to server
        \STATE \quad Compute public logits $\phi_{i}^{(t)}$ on $\mathcal{D}_{\text{pub}}$ and upload to server
    \ENDFOR
    \STATE Server aggregates logits to compute collaborative loss using weights $\{W_i^{(t)}\}$ from $F_{i}^{(t)}$
    \FOR{each client $i$}
        \STATE \textbf{Phase 2 (Collaborative training)}: Train on $\mathcal{D}_{\text{pub}}$ using collaborative loss
        \STATE \textbf{Phase 3 (Private retraining)}:
        \FOR{each local epoch $e = 1, \dots, E$}
            \STATE Perform DLR on $\mathcal{D}_i$
            \STATE Train on $\mathcal{D}_i$ using SL loss with DLR soft labels
        \ENDFOR
    \ENDFOR
\ENDFOR
\end{algorithmic}
\end{algorithm}

\subsubsection*{CCR vs. ECCR}
\hspace{1.5em}Client Confidence Re-weighting (CCR) was proposed by the RHFL+ research group in their earlier work~\cite{fang2022robust}. The key difference between CCR and ECCR lies in the confidence formulation: while CCR considers only the drop in SL loss to estimate a client’s learning efficiency, ECCR additionally incorporates the normalized magnitude of model updates. The motivation behind ECCR is that clients with larger parameter updates are more likely to have local data distributions that deviate from the global public knowledge, and should therefore be assigned lower confidence during aggregation. However, our ablation study suggests that, contrary to this theoretical expectation, CCR empirically outperforms ECCR in most scenarios. We hypothesize that this is because ECCR may over-penalize clients when update magnitudes are unstable. 

\subsection{NVFlare}
\hspace{1.5em}NVFlare is an open-source FL framework developed by NVIDIA. It uses a custom communication protocol built upon gRPC over TCP/IP, which is secure, reliable and efficient. SSL/TLS can be configured to impose encryption for data confidentiality.

\begin{figure}[h]
    \centering
    \includegraphics[width=0.7\textwidth]{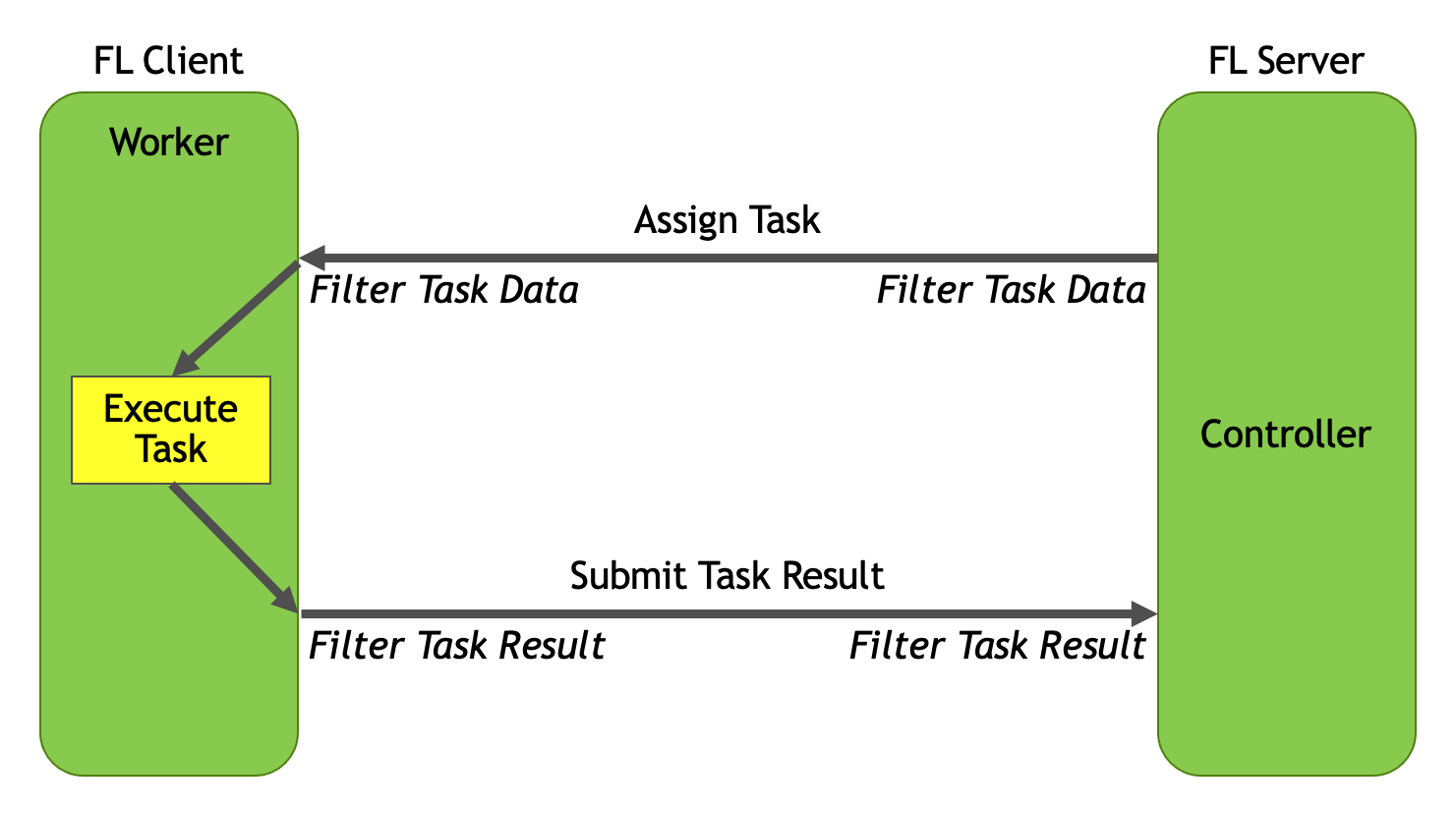}
    \caption{NVFlare Learner-Controller Workflow, adapted from\cite{nvflare}}
    \label{fig:controller_worker_flow}
\end{figure}

One of the primary advantages of NVFlare is its modular, plugin-based architecture, which enables users to define separate components such as Learner and Controller. This separation of concerns facilitates the implementation of new algorithms. Figure \ref{fig:controller_worker_flow} shows the overall workflow of an NVFlare training process. In addition, NVFlare is designed with fault tolerance. It includes mechanisms such as health checks and timeout recovery, which make it resilient to network disruptions and unstable client behavior. Moreover, NVFlare stands out from many research-oriented FL frameworks, such as Flower~\cite{flowerFlowerFramework}, through its production-ready architecture. It provides built-in support for secure job provisioning, role-based access control and scalable deployment, making it suitable for real-world enterprise applications.

NVFlare's simulator mode is recommended to be used for the research experiments. As stated in the official documentation \cite{nvflareNVIDIAFLARE}, this mode closely mimics real-world deployment while allowing experiments to be conducted on a single node. This is particularly suitable for running GPU-based proof-of-concept experiments on the CSD3 cluster without requiring long-term multi-node reservations. Moreover, the code can be directly used in real-life deployment.

\subsection{Dataset}

\hspace{1.5em}The private dataset used in all experiments is CIFAR-10\cite{Krizhevsky09learningmultiple}, while CIFAR-100\cite{Krizhevsky09learningmultiple} serves as the public auxiliary dataset, in line with the experimental setup of the original RHFL+ paper. CIFAR-10 comprises 60000 $32\times32$ RGB images across 10 classes, whereas CIFAR-100 contains images of the same resolution but from 100 fine-grained categories, making it well-suited as a diverse, unlabeled proxy for public knowledge transfer.

To ensure experimental fairness and reproducibility, we adopted the official RHFL+ codebase for data partitioning and label noise injection, thereby isolating the algorithmic component as the only variable across different methods.

It is important to note that the original RHFL+ implementation introduces label noise into both private and public datasets. However, this deviates from the theoretical formulation of RHFL+, which assumes that the public dataset remains clean to facilitate effective collaborative learning. Injecting noise into the public dataset may degrade the performance of knowledge distillation, misguide the global prototype alignment and ultimately diminish the algorithm's robustness to private label corruption. Accordingly, in my experiments, I maintain a noise-free public dataset and inject noise solely into the private datasets held by clients, preserving the intended separation of clean supervision and noisy local labels.

\subsection{Noise Types}

\hspace{1.5em}To simulate real-world data imbalances in annotation quality, we inject synthetic label noise into the clients' local CIFAR-10 datasets. Two widely used label corruption strategies are considered, capturing different types of noise behavior. The first type is symmetric noise, where each label is independently flipped to one of the remaining $C-1$ classes with uniform probability $\mu$. This simulates random annotation errors without semantic bias. The other type is pairflip noise, where labels are flipped to a semantically adjacent class according to a deterministic permutation, introducing more structured mislabeling, which may arise in fine-grained classification tasks. Following the original RHFL+ setup, noise rates $\mu \in \{0.1, 0.2\}$ are explored to reflect mild and moderate noise levels. 

\subsection{Evaluation Metrics}

\hspace{1.5em}The primary evaluation metric is classification accuracy on the clean CIFAR-10 test set. Accuracy is a widely accepted measure for multi-class image classification tasks, providing an intuitive indicator of model performance. However, Accuracy may fail to fully capture performance in imbalanced or domain-specific settings, such as medical imaging or anomaly detection.

In future extensions to medical imaging datasets, additional evaluation metrics such as area under the curve (AUC) are used to complement accuracy, particularly in cases of label imperfections.

\subsection{Experimental Setup}

\hspace{1.5em}To rigorously evaluate the robustness and effectiveness of RHFL+, we reproduce both the ablation studies and the comparative experiments against the state-of-the-art methods. Following the original paper, we adopt a federated learning simulation with multiple heterogeneous clients and one central server.

First, to assess the contribution of each component in the RHFL+ pipeline, we perform an ablation study by selectively disabling individual modules. These include the base Heterogeneous Federated Learning (HFL), the Symmetric Cross Entropy Learning (SL) loss, the Dynamic Label Refinement (DLR) and the Enhanced Client Confidence Re-Weighting (ECCR). This setup enables us to quantify the impact of each component on overall performance. 

Then, for comparative evaluation, RHFL+ is benchmarked against a diverse set of recent and representative FL algorithms, including:
\textit{FedMD, FedDF, KT-pFL, FedProto, FCCL, FedGH, FedTGP, RHFL and AugHFL} (where applicable). Each baseline is implemented and adapted to a unified framework for consistent evaluation.

\vspace{1em}
\begin{table}[ht]
\centering
\begin{tabular}{ll}
\toprule
\textbf{Parameter} & \textbf{Value} \\
\midrule
ECCR impact of confidence $\eta$ & 1.2 \\
Loss function & Symmetric Cross Entropy (SCE) \\
Learning rate & 0.001 \\
SL parameters & $\lambda = 0.4$,  $\gamma = 0.9$ \\
Temperature & 4.0 \\
collaborative learning epochs & 40 \\
Client models & ResNet10, ResNet12, ShuffleNet, MobileNetV2 \\
Public dataset & CIFAR-100  \\
DLR maximum influence of predictions & 10.0 \\
\bottomrule
\end{tabular}
\caption{RHFL+ Configuration}
\label{tab:rhfl_config}
\end{table}

The RHFL+ configuration is aligned with the hyperparameter settings reported in the original paper, which is summarized in the Table \ref{tab:rhfl_config}.

\subsection{Challenges}
During implementation, the following challenges were encountered:
\paragraph{FCCL Memory Challenge} The FCCL algorithm computes a cross-correlation matrix between each client’s logits and the averaged global logits on the public dataset (Equation~\ref{eq:fccl_corr}). This requires broadcasting and storing high-dimensional logits across all clients and public samples, which led to frequent CUDA out-of-memory errors in our distributed setup. To address this, we implemented a batch-wise computation strategy to reduce peak memory usage. However, this approximation may have introduced numerical inconsistencies or weakened the collaborative signal, potentially contributing to the performance gap observed relative to the original paper.

    \begin{equation}
    \mathcal{M}^{uv}_i \triangleq 
    \frac{
        \sum_b \left\| Z^{b,u}_i \right\| \cdot \left\| \bar{Z}^{b,v} \right\|
    }{
        \sqrt{ \sum_b \left\| Z^{b,u}_i \right\|^2 } \cdot \sqrt{ \sum_b \left\| \bar{Z}^{b,v} \right\|^2 }
    },
    \label{eq:fccl_corr}
    \end{equation}

\paragraph{AugHFL} While AugHFL\cite{10378155} is included in the main comparison table of the RHFL+ paper, we chose to exclude it from our experimental evaluation due to a fundamental mismatch in the types of corruption assumptions. RHFL+ is designed to address \emph{semantic label noise}, where the image content remains clean but the associated labels may be incorrect (e.g. symmetric or pairflip noise). In contrast, AugHFL specifically targets \emph{input-level corruption}, such as visual distortions (e.g. blur, fog, noise), under the assumption of clean labels. As the two methods are optimized for fundamentally different types of data corruption, direct comparison may lead to misleading conclusions. Additionally, comparison methodology of AugHFL with RHFL+ is not disclosed in the paper. Our focus is on evaluating label-noise-robust federated learning methods under the same corruption. Therefore, we restrict comparisons to methods that address noisy-label scenarios, which aligns with the intended scope of RHFL+. 

\paragraph{Complexity of Experimental Configuration} Given the scope of this study: spanning over 10 federated learning methods, 5 noise rates, 2 noise types and 4 different private datasets, the experimental setup is inherently complex and infeasible to manage manually. To address this, I designed a modular configuration system using structured config files that can be directly accessed by both the controller and learner components in NVFlare. This system enables a generic and plugin-friendly codebase, making it straightforward to integrate new algorithms and automate large-scale experimentation. A script can iterate over all experimental dimensions simply by modifying the relevant configuration files. Specifically, a base configuration file defines dataset-invariant parameters (e.g. seed, loss functions), while dataset-specific files define parameters such as private data length and batch size. This design significantly reduces repetitive code and ensures reproducibility across experiments.

\paragraph{Custom Components} RHFL+ does not use a central model, unlike traditional FL algorithms. Instead, it relies on knowledge distillation via public data and supports heterogeneous client models, which breaks the assumption of NVFlare’s default controllers like ScatterAndGather\cite{nvflareScatterGather} that require a shared global model. As a result, these built-in workflows cannot be used directly. To support both RHFL+ (no central model) and standard algorithms (with a central model) in the same framework, we implemented custom Controller\cite{nvflareControllersController} and Executor components in NVFlare to coordinate the distillation process and allow flexible integration of both paradigms.

\subsection{Pipeline}
\begin{figure}[H]
    \centering
    \includegraphics[width=\textwidth]{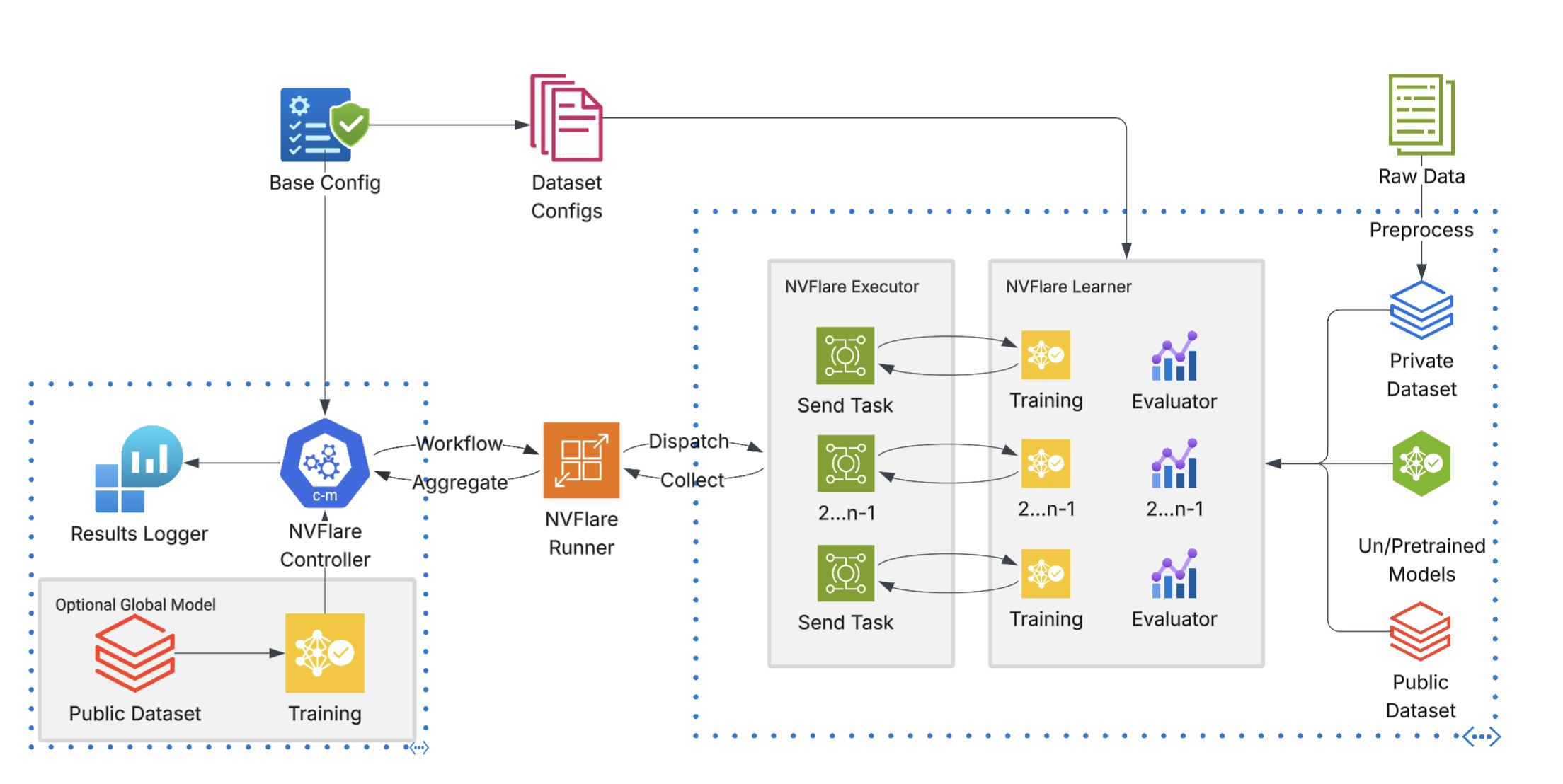}
    \caption{Data Pipeline of the Experiments Using NVFlare Framework}
    \label{fig:pipeline}
\end{figure}

\hspace{1.5em}Before discussing the results, Figure \ref{fig:pipeline} provides a high-level overview of the pipeline. The bottom-left section represents the Server, while the bottom-right section corresponds to the Clients. The NVFlare Runner orchestrates communication between the Controller on the server side and the Learners on the client side. Depending on the specific federated learning algorithm, the server may perform global model training. Since our evaluation focuses on the client models, the evaluation process is encapsulated as a task executed on the client side and subsequently sent back to the Controller for final formatting and logging.

\subsection{Results}
\hspace{1.5em}The results of the RHFL+ ablation study under different label noise conditions are summarized in Tables~\ref{tab:rate01_ablation}--\ref{tab:eccr_ablation}. The experiments assess the contribution of individual components by progressively enabling them and measuring their effect on average client performance across four heterogeneous models ($\theta_1$ to $\theta_4$). We report results for both symmetric and pairflip noise types, at noise levels $\mu = 0.1$ and $\mu = 0.2$.

\begin{table}[htbp]
\centering
\resizebox{\textwidth}{!}{%
\begin{tabular}{cccc|cccccc|ccccc}
\toprule
\multicolumn{4}{c|}{\textbf{Components}} & \multicolumn{6}{c|}{\textbf{Pairflip}} & \multicolumn{5}{c}{\textbf{Symflip}} \\
\textbf{HFL} & \textbf{SL} & \textbf{DLR} & \textbf{ECCR} & $\theta_1$ & $\theta_2$ & $\theta_3$ & $\theta_4$ & \textbf{Avg} & & $\theta_1$ & $\theta_2$ & $\theta_3$ & $\theta_4$ & \textbf{Avg} \\
\midrule
& & & & 79.06 & 77.66 & 66.63 & 74.02 & 74.34 & & 79.54 & 79.04 & 65.23 & 74.54 & 74.54 \\
\checkmark & & & & 79.81 & 78.14 & 73.69 & \textbf{80.64} & 78.07 & & 75.39 & 77.52 & 71.97 & 78.64 & 75.88 \\
& \checkmark & & & 79.50 & 78.99 & 71.77 & 79.24 & 77.38 & & 79.25 & 81.61 & 71.08 & 79.00 & 77.74 \\
\checkmark & \checkmark & & & 80.26 & 81.88 & 74.31 & 80.51 & 79.24 & & 79.25 & 82.85 & \textbf{73.04} & \textbf{80.74} & 78.97 \\
\checkmark & \checkmark & \checkmark & & 81.98 & \textbf{82.67} & \textbf{74.53} & 78.88 & 79.52 & & 82.93 & 61.43 & 71.34 & 79.01 & 73.67 \\
\checkmark & \checkmark & \checkmark & \checkmark & \textbf{83.00} & 82.00 & 74.32 & 80.10 & \textbf{79.86} & & \textbf{83.73} & \textbf{83.09} & 71.46 & 78.53 & \textbf{79.20} \\
\bottomrule
\end{tabular}
}
\caption{Ablation Study with the Noise Rate $\mu = 0.1$, $\theta_k$ represents the Local Model of the Client $c_k$, CIFAR-10 as the Private Dataset, CIFAR-100 as the Public Dataset}
\label{tab:rate01_ablation}
\end{table}

Table~\ref{tab:rate01_ablation} shows that at $\mu = 0.1$, the full RHFL+ configuration (all components enabled) consistently yields the highest average accuracy for both noise types. Specifically, utilizing HFL results in substantial gains, highlighting the importance of refined pseudo-labeling. ECCR further improves results slightly in most cases. However, in Table~\ref{tab:rate02_ablation} (for $\mu = 0.2$), the best performance under pairflip noise is observed when ECCR is omitted, suggesting that ECCR may be less robust under severe structured noise conditions.

In Table~\ref{tab:eccr_ablation}, we isolate the effect of reweighting strategies. Both CCR and ECCR improve over the baseline with no reweighting. Interestingly, CCR slightly outperforms ECCR in 3 out of 4 conditions, particularly under higher noise levels. This suggests that simpler CCR may generalize better across clients than more complex ECCR mechanisms.

\paragraph{Analysis} Overall, the results support the design of RHFL+ as a robust approach from data imperfections. The combination of SL and DLR plays a vital role in improving model generalization under label noise, while the HFL backbone ensures adaptability to heterogeneous architectures. Although ECCR is effective in many cases, it does not outperform CCR in most cases, implying that confidence-based aggregation strategies might be unstable. In conclusion, HFL+SL+DLR+CCR appears to be the best-performing combination.

\begin{table}[htbp]
\centering
\resizebox{\textwidth}{!}{%
\begin{tabular}{cccc|ccccc|ccccc}
\toprule
\multicolumn{4}{c|}{\textbf{Components}} & \multicolumn{5}{c|}{\textbf{Pairflip}} & \multicolumn{5}{c}{\textbf{Symflip}} \\
\textbf{HFL} & \textbf{SL} & \textbf{DLR} & \textbf{ECCR} & $\theta_1$ & $\theta_2$ & $\theta_3$ & $\theta_4$ & \textbf{Avg} & $\theta_1$ & $\theta_2$ & $\theta_3$ & $\theta_4$ & \textbf{Avg} \\
\midrule
& & & & 75.86 & 76.70 & 64.25 & 73.11 & 72.48 & 77.46 & 77.15 & 64.36 & 71.52 & 72.51 \\
\checkmark & & & & 73.08 & 72.88 & 70.00 & \textbf{78.34} & 73.58 &  70.70 & 69.23 & 67.34 & 74.40 & 70.42 \\
& \checkmark & & & 71.25 & 68.94 & 66.48 & 77.32 & 71.00 & 75.09 & 74.01 & 70.56 & 76.24 & 73.98 \\
\checkmark & \checkmark & & & 76.28 & 75.74 & \textbf{72.18} & 77.78 & 75.50 & 78.19 & 80.26 & \textbf{71.00} & \textbf{77.70} & 76.79 \\
\checkmark & \checkmark & \checkmark & & \textbf{80.77} & \textbf{81.73} & 69.93 & 76.50 & \textbf{77.23} & 81.39 & 80.84 & 70.12 & 77.17& 77.38 \\
\checkmark & \checkmark & \checkmark & \checkmark & 80.02 & 81.07 & 69.95 & 76.90 & 76.98 & \textbf{82.29} & \textbf{81.80} & 70.39 & 77.24 & \textbf{77.93} \\
\bottomrule
\end{tabular}
}
\caption{Ablation Study with the Noise Rate $\mu = 0.2$, $\theta_k$ represents the Local Model of the Client $c_k$, CIFAR-10 as the Private Dataset, CIFAR-100 as the Public Dataset}
\label{tab:rate02_ablation}
\end{table}

\begin{table}[htbp]
\centering
\begin{tabular}{lcc|cc}
\toprule
\textbf{Method} & \multicolumn{2}{c|}{$\mu = 0.1$} & \multicolumn{2}{c}{$\mu = 0.2$} \\
 & \textbf{Pairflip} & \textbf{Symflip} & \textbf{Pairflip} & \textbf{Symflip} \\
\midrule
w/o Reweighting & 79.52 & 73.68 & 77.23 & 77.38 \\
+CCR            & 79.75 & \textbf{79.62} & \textbf{77.57} & \textbf{77.98} \\
+ECCR           & \textbf{79.86} & 79.20 & 76.99 & 77.93 \\
\bottomrule
\end{tabular}
\caption{Ablation Study of ECCR, $\mu$ denotes the noise rate, CIFAR-10 as the Private Dataset, CIFAR-100 as the Public Dataset}
\label{tab:eccr_ablation}
\end{table}

As shown in Table~\ref{tab:main_01} and Table~\ref{tab:main_02}, which show the comparison of all state-of-art FL methods, RHFL+ outperforms all baseline methods in most of the cases, across both pairflip and symmetric noise settings, under noise rates $\mu = 0.1$ and $\mu = 0.2$. Results are reported across four heterogeneous private models ($\theta_1$ to $\theta_4$), with the average accuracy used for final comparison.

At $\mu = 0.1$, RHFL+ achieves average accuracy of 79.86\% on pairflip noise and 79.20\% on symmetric noise. This outperforms or perform similarly to all other baselines, including KT-pFL (79.25\%, 78.68\%) and RHFL (78.34\%, 79.22\%), indicating RHFL+’s superior handling of moderately corrupted labels. RHFL+ also shows good performance consistency across all client models, particularly improving weaker models such as $\theta_3$ compared to others.

At a higher noise level ($\mu = 0.2$), the performance gap widens. RHFL+ maintains strong robustness with average accuracies of 76.98\% (pairflip) and 77.93\% (symflip), whereas FedMD drops to 73.35\% and 71.39\% and FedDF collapses below 66\% in all cases. 


\paragraph{Analysis}

Overall, RHFL+ offers the most stable and effective performance in noisy and heterogeneous federated settings. Its modular combination of HFL, SL, DLR and ECCR not only boosts average performance but also mitigates underperformance in weaker client models. While other methods without dynamic label correction may fail in high-noise federated setups.

\begin{table}[htbp]
\centering
\resizebox{\textwidth}{!}{%
\begin{tabular}{lcccccc|ccccc}
\toprule
\multicolumn{1}{c}{} & \multicolumn{6}{c|}{\textbf{Pairflip}} & \multicolumn{5}{c}{\textbf{Symflip}} \\
\textbf{Method} & $\theta_1$ & $\theta_2$ & $\theta_3$ & $\theta_4$ & \textbf{Avg} & & $\theta_1$ & $\theta_2$ & $\theta_3$ & $\theta_4$ & \textbf{Avg} \\
\midrule
Baseline    & 79.06 & 77.66 & 66.63 & 74.02 & 74.34 & & 79.54 & 79.04 & 65.23 & 74.54 & 74.54 \\
FedMD\cite{li2019fedmdheterogenousfederatedlearning}  & 79.84 & 77.91 & 71.68 & 80.36 & 77.45 & & 77.54 & 78.14 & 71.03 & 77.69 & 76.10 \\
FedDF\cite{lin2020ensemble}  & 76.74 & 75.48 & 67.89 & 75.74 & 73.78 & & 73.15 & 73.10 & 64.71 & 74.01 & 71.24 \\
KT-pFL\cite{zhang2021parameterizedknowledgetransferpersonalized} & 81.04 & 81.30 & 74.08 & \textbf{80.58} & 79.25 & & 80.94 & 81.23 & \textbf{72.57} & 79.96 & 78.68 \\
FedProto\cite{tan2022fedprotofederatedprototypelearning}    & 74.22 & 74.41 & 68.12 & 78.43 & 73.80 & & 72.49 & 71.50 & 64.57 & 71.80 & 70.09 \\
FCCL\cite{9879190}   & 72.82 & 73.55 & 58.83 & 73.03 & 69.56 & & 73.09 & 71.68 & 56.19 & 69.94 & 67.73 \\
FedGH\cite{yi2023fedghheterogeneousfederatedlearning}  & 76.22 & 78.60 & 68.18 & 79.81 & 75.70 & & 74.45 & 74.50 & 66.99 & 78.02 & 73.49 \\
FedTGP\cite{zhang2024fedtgptrainableglobalprototypes} & 76.22 & 74.79 & 65.78 & 78.67 & 73.87 & & 74.77 & 74.93 & 65.22 & 75.30 & 72.56 \\
RHFL\cite{fang2022robust}    & 81.75 & 79.71 & 72.64 & 79.25 & 78.34 & & 81.97 & 82.85 & 71.44 & \textbf{80.63} & \textbf{79.22} \\
\textbf{RHFL+}\cite{10816157} & \textbf{83.00} & \textbf{82.00} & \textbf{74.32} & 80.10 & \textbf{79.86} & & \textbf{83.73} & \textbf{83.09} & 71.46 & 78.53 & 79.20 \\
\bottomrule
\end{tabular}
}
\caption{CIFAR-10 as the Private Dataset, CIFAR-100 as the Public Dataset, with the Noise Rate $\mu = 0.1$}
\label{tab:main_01}
\end{table}

\begin{table}[htbp]
\centering
\resizebox{\textwidth}{!}{%
\begin{tabular}{lcccccc|ccccc}
\toprule
\multicolumn{1}{c}{} & \multicolumn{6}{c|}{\textbf{Pairflip}} & \multicolumn{5}{c}{\textbf{Symflip}} \\
\textbf{Method} & $\theta_1$ & $\theta_2$ & $\theta_3$ & $\theta_4$ & \textbf{Avg} & & $\theta_1$ & $\theta_2$ & $\theta_3$ & $\theta_4$ & \textbf{Avg} \\
\midrule
Baseline    & 75.86 & 76.70 & 64.25 & 73.11 & 72.48 && 77.46 & 77.15 & 64.36 & 71.52 & 72.51 \\
FedMD\cite{li2019fedmdheterogenousfederatedlearning}  & 72.74 & 73.18 & 68.78 & \textbf{78.71} & 73.35 & & 71.91 & 69.98 & 69.37 & 74.28 & 71.39 \\
FedDF\cite{lin2020ensemble}  & 69.69 & 66.98 & 59.77 & 66.14 & 65.65 & & 65.32 & 65.23 & 57.05 & 65.56 & 63.29 \\
KT-pFL\cite{zhang2021parameterizedknowledgetransferpersonalized} & 74.00 & 76.20 & 67.66 & 76.17 & 74.00 & & 76.21 & 76.64 & 68.79 & 73.70 & 73.84 \\
FedProto\cite{tan2022fedprotofederatedprototypelearning}    & 71.61 & 66.60 & 64.01 & 76.93 & 69.79 & & 70.75 & 74.25 & 66.03 & 68.63 & 69.92 \\
FCCL\cite{9879190}   & 72.41 & 70.75 & 57.21 & 70.11 & 67.62 & & 69.66 & 69.05 & 60.82 & 70.60 & 67.53 \\
FedGH\cite{yi2023fedghheterogeneousfederatedlearning}  & 66.71 & 72.44 & 63.96 & 77.72 & 70.21 & & 67.62 & 69.49 & 63.44 & 73.80 & 68.59 \\
FedTGP\cite{zhang2024fedtgptrainableglobalprototypes} & 70.39 & 64.61 & \textbf{75.80} & 59.61 & 67.60 & & 68.65 & 65.14 & 61.84 & 72.22 & 66.96 \\
RHFL\cite{fang2022robust}    & 72.47 & 77.24 & 71.00 & 77.36 & 74.52 & & 77.78 & 79.53 & \textbf{71.54} & 76.84 & 76.42 \\
\textbf{RHFL+}\cite{10816157} & \textbf{80.02} & \textbf{81.07} & 69.95 & 76.90 & \textbf{76.98} && \textbf{82.29} & \textbf{81.80} & 70.39 & \textbf{77.24} & \textbf{77.93} \\
\bottomrule
\end{tabular}
}
\caption{CIFAR-10 as the Private Dataset, CIFAR-100 as the Public Dataset, with the Noise Rate $\mu = 0.2$}
\label{tab:main_02}
\end{table}

\begin{figure}[H]
    \centering
    \begin{subfigure}[t]{\textwidth}
        \centering
        \includegraphics[width=\textwidth]{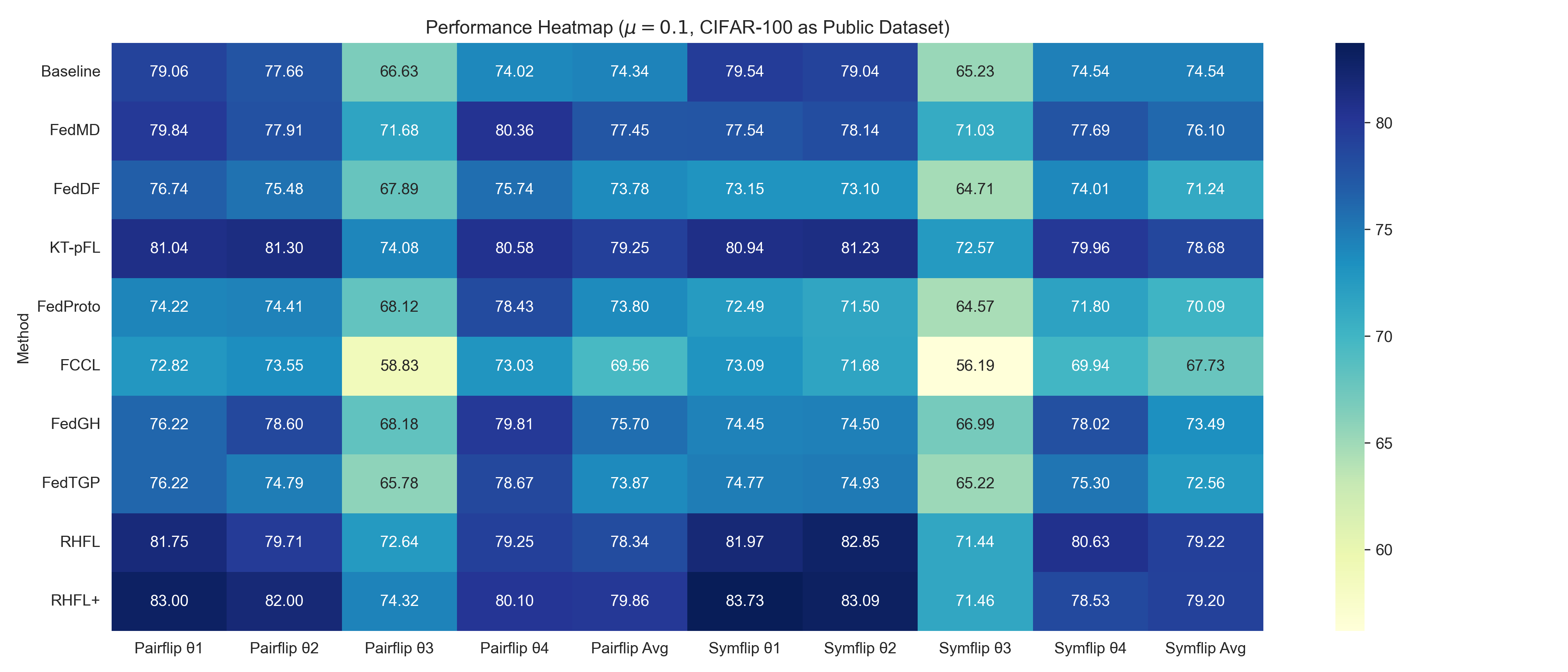}
        \caption{Performance Heatmap with $\mu=0.1$}
        \label{fig:heatmap_mu_01}
    \end{subfigure}

    \vspace{1em} 

    \begin{subfigure}[t]{\textwidth}
        \centering
        \includegraphics[width=\textwidth]{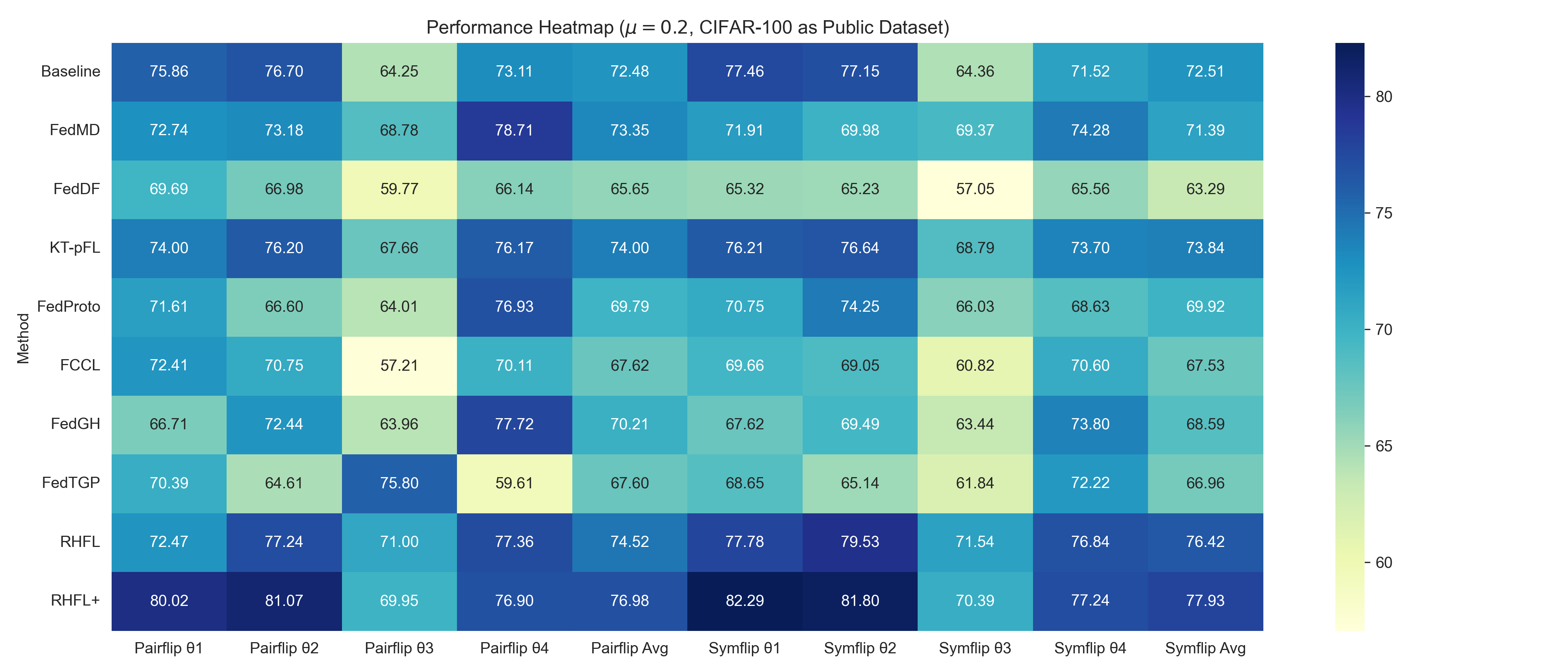}
        \caption{Performance Heatmap with $\mu=0.2$}
        \label{fig:heatmap_mu_02}
    \end{subfigure}
    
    \caption{Comparison of all methods performance under different noise levels ($\mu=0.1$ vs $\mu=0.2$)}
    \label{fig:heatmap_comparison_vertical}
\end{figure}

\begin{figure}[h]
    \centering
    \begin{subfigure}[t]{0.48\textwidth}
        \centering
        \includegraphics[width=\textwidth]{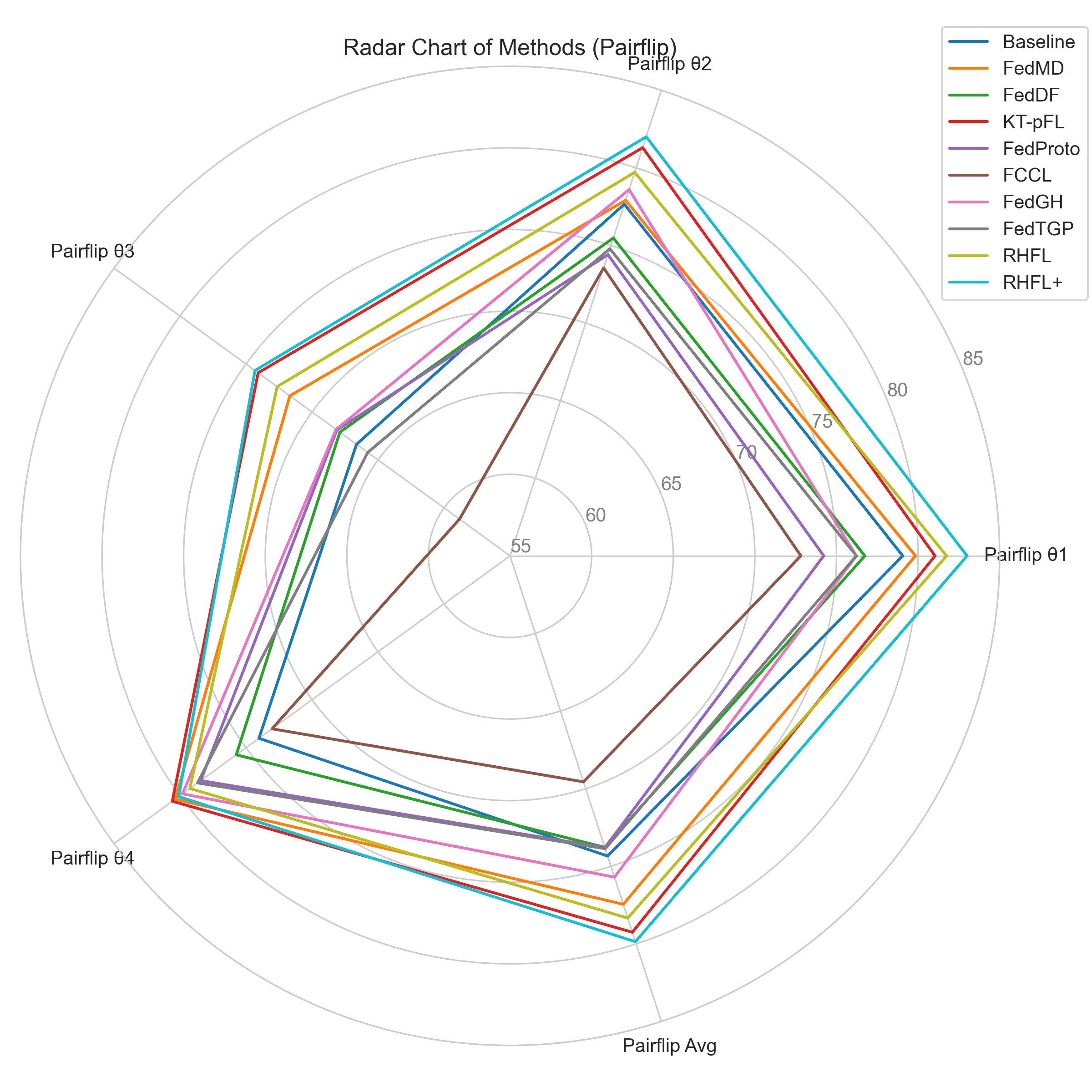}
        \caption{Performance Radar Chart with $\mu=0.1$}
        \label{fig:radar_mu_01}
    \end{subfigure}
    \hfill
    \begin{subfigure}[t]{0.48\textwidth}
        \centering
        \includegraphics[width=\textwidth]{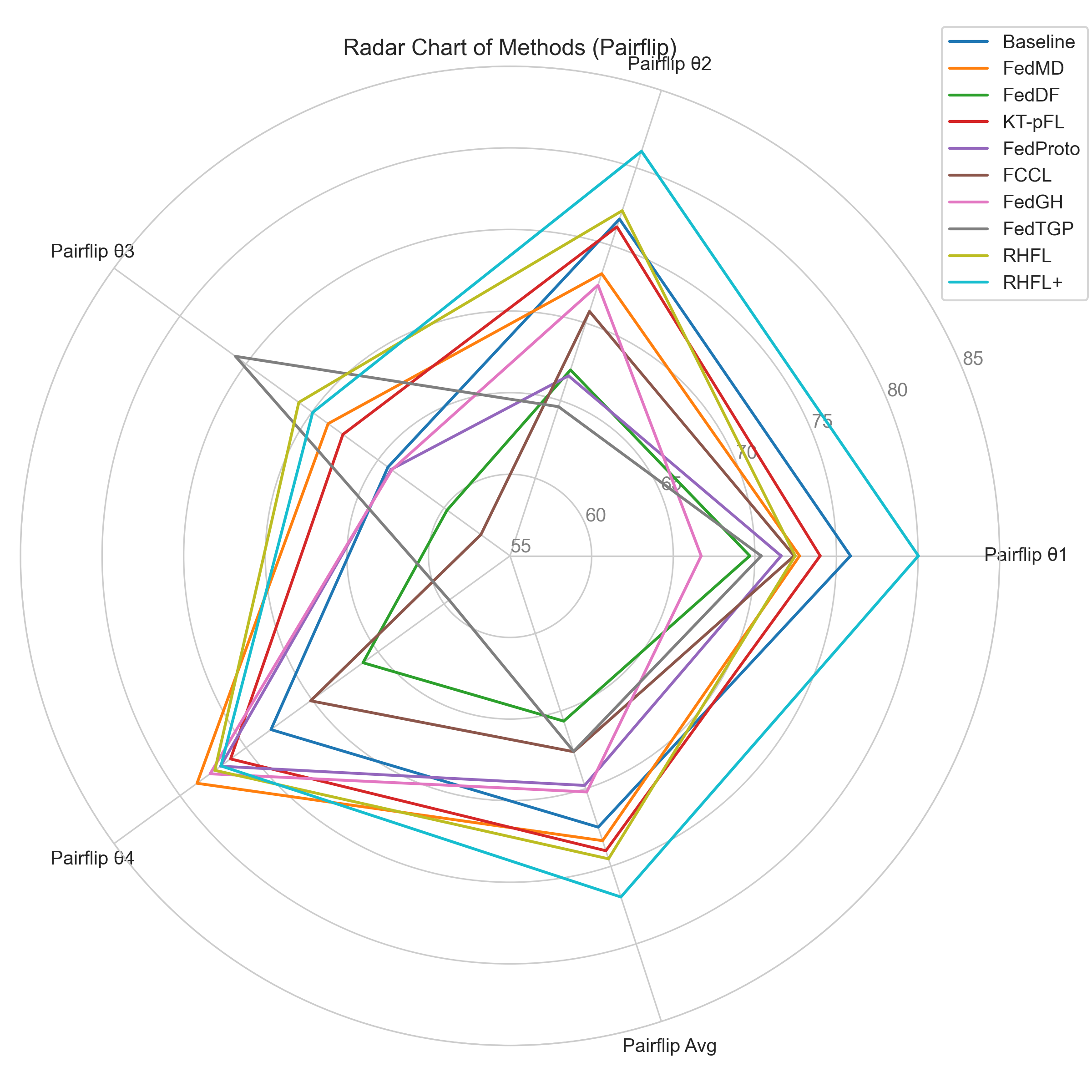}
        \caption{Performance Radar Chart with $\mu=0.2$}
        \label{fig:radar_mu_02}
    \end{subfigure}
    \caption{Comparison of all methods performance under different noise levels ($\mu=0.1$ vs $\mu=0.2$)}
    \label{fig:radar_comparison}
\end{figure}

\paragraph{Visualization}
With more direct visualization provided in Figures~\ref{fig:heatmap_comparison_vertical} and~\ref{fig:radar_comparison}, we observe that RHFL-based methods (RHFL and RHFL+) consistently outperform other algorithms under both low and high noise conditions. 

Most baseline methods show degraded performance under increasing noise, with some performing comparably at low noise levels but dropping significantly under higher noise. The radar charts clearly highlight this trend: while RHFL+ maintains robust performance across all pairflip noise configurations, other methods exhibit inconsistency.

These results indicate that RHFL-based approaches are not only more accurate on average but also more resilient to varying types and degrees of label noise.

\subsection{Original vs Reproduced}
\hspace{1.5em}According to the plots of original paper, RHFL+ with ECCR consistently outperforms all other methods. However, in our reproduced experiments, we observe a notable deviation: CCR frequently achieves the highest average accuracy, particularly under the pairflip setting. Despite this, the accuracy of both CCR and ECCR in our experiments exceeds the original reported values. This improvement may be attributed to differences in random seed initialization or a more faithful implementation of the theoretical pipeline.

Furthermore, baseline method performance is similar in our reproduction. A notable exception is FCCL, which consistently lags behind the original results. As discussed earlier, our batch-wise adaptation to mitigate CUDA memory overflow likely introduced approximation errors in the computation of the cross-correlation matrix, leading to reduced performance.

\paragraph{CCR vs. ECCR Analysis}
While both CCR and ECCR apply confidence-based reweighting, ECCR introduces a normalization term that scales weights based on entropy, as shown in Equation\ref{eq:le}. In practice, this normalization might diminish the relative contributions of high-confidence clients by compressing the weight range, particularly when the entropy values across clients are not sharply differentiated. By contrast, CCR maintains a raw confidence-driven reweighting strategy, which may result in a more discriminative client selection, especially under heterogeneous noise conditions. This may explain why CCR yields better average performance.

\section{Further Research}
\label{sec:extension}

\subsection{Scaling Test}

\hspace{1.5em}Before diving into the medical imaging domain using FL, a scaling experiment was first conducted on CIFAR-10 to evaluate the scalability of my codebase, as well as the overall effectiveness of the FL method in a larger scale.

Since the codebase reads from a centralized configuration file, scaling the number of clients was straightforward. For consistency, I used the same model: ResNet12, on all clients. The main challenge was acquiring sufficient GPU resources on CSD3. To simplify resource management, I requested four GPUs per job (the maximum allowed per submission), thus utilizing the full capacity of a GPU node. Clients were assigned to GPUs using modulo-based allocation to ensure an even distribution of computation load. This strategy allowed me to simulate up to 100 clients on a single node by leveraging multithreading effectively.

\begin{figure}[H]
    \centering
    \includegraphics[width=0.7\textwidth]{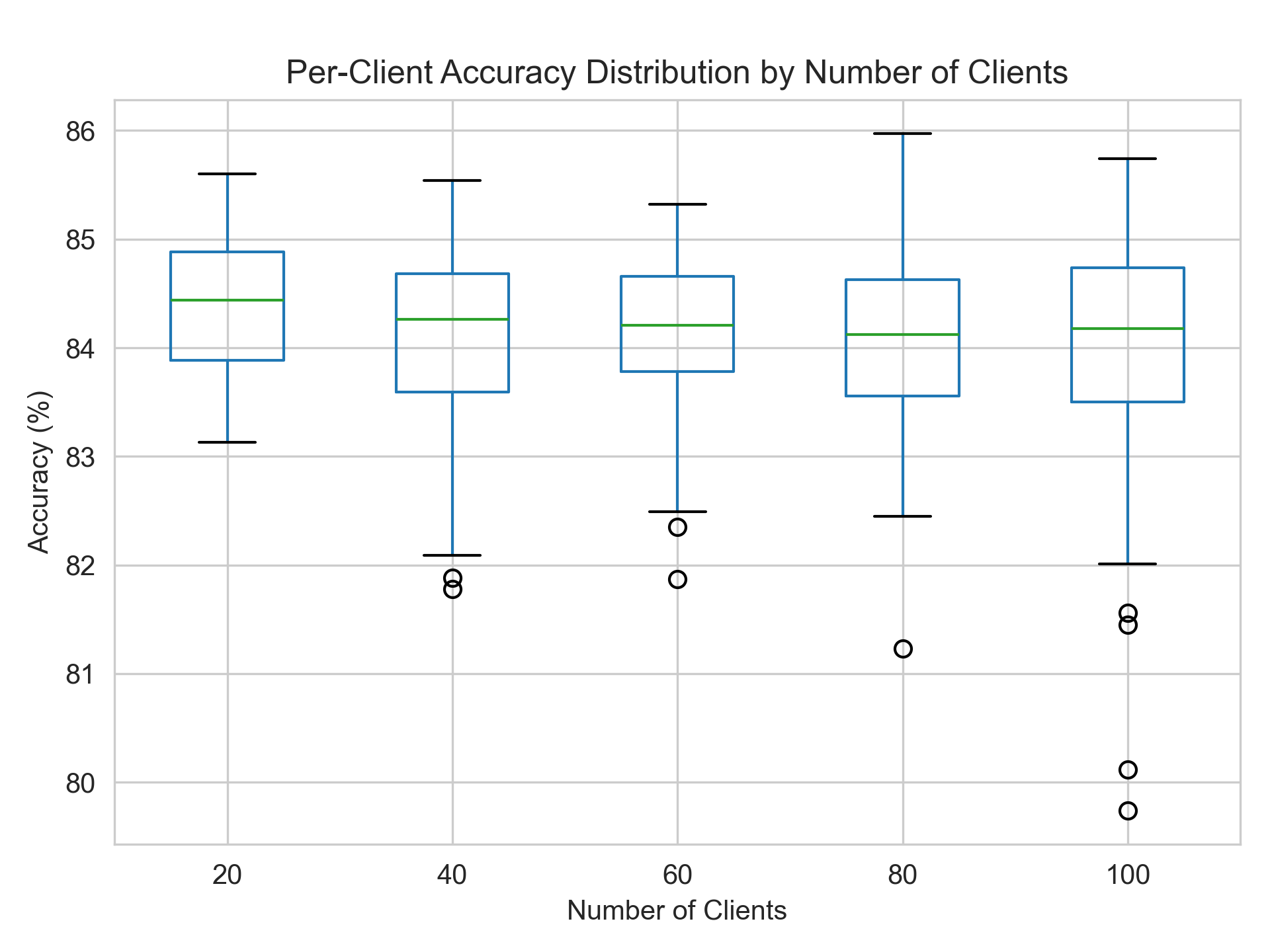}
    \caption{Test accuracy of RHFL+ with increasing number of clients, CIFAR-10 as the Private Dataset, CIFAR-100 as the Public Dataset}
    \label{fig:scale_accuracy}
\end{figure}

Figure~\ref{fig:scale_accuracy} illustrates the per-client accuracy distribution across different numbers of clients. As observed, the performance remains relatively stable as the number of clients increases, with only minor fluctuations. This demonstrates the robustness of both the RHFL+ algorithm and my FL code.

\subsection{Random Noise Rate Test}
\hspace{1.5em}In the current setup, all clients experience the same noise rate, which does not fully reflect the data heterogeneity often observed in real-world federated learning scenarios. As a next step, we simulate varying noise levels across clients to evaluate how RHFL+ estimates and utilizes client confidence under more realistic, non-i.i.d. conditions.

To simulate this, we assigned each of the 10 clients a random symmetric noise rate uniformly sampled from the range [0.0, 0.5]. We then evaluated the performance of RHFL+ (either CCR or ECCR) against a non-collaborative baseline (LocalOnly) over 40 communication rounds.

\begin{figure}[h]
    \centering
    \includegraphics[width=\textwidth]{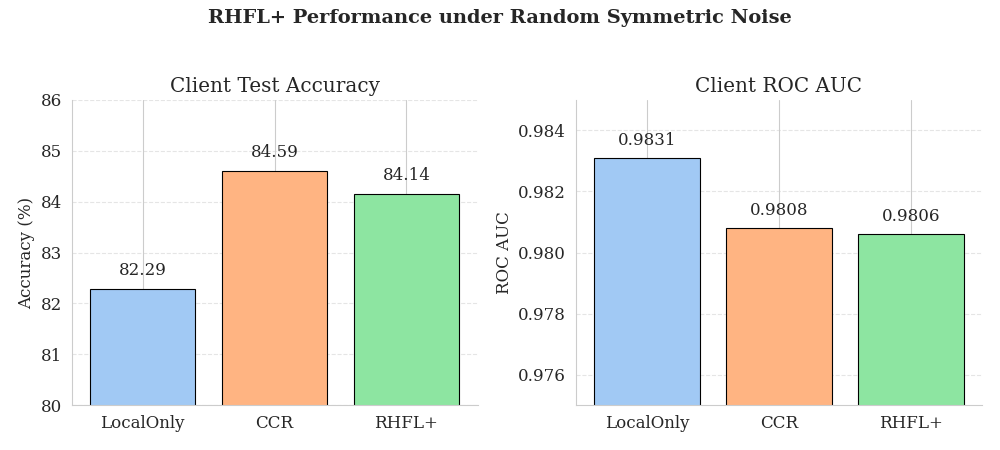}
    \caption{Test accuracy (Left) and AUC (right) of three methods with random client noise rates, CIFAR-10 as the Private Dataset, CIFAR-100 as the Public Dataset}
    \label{fig:random_noise}
\end{figure}

Figure~\ref{fig:random_noise} shows the final averaged test accuracy and ROC AUC across all clients. RHFL-based methods clearly outperform the local-only baseline by over 2\% in accuracy on average. This demonstrates the strong robustness of RHFL+ in noise heterogeneous environments, especially under non-IID data distributions and random label corruption.

The results highlight RHFL+'s ability to leverage cross-client collaboration to mitigate the impact of unreliable labels, achieving both higher accuracy and similar AUC performance.

\subsection{FL for Medical Imaging}
\hspace{1.5em}In the medical domain, such as hospital environments, data is often distributed across multiple institutions, where data is typically limited in size, imbalanced and expensive to collect and label due to the need for expert knowledge~\cite{diagnostics13193140}. Small datasets make it challenging to train effective models, as they can lead to biased performance estimates and overfitting~\cite{vabalas_machine_2019}. Moreover, when class distributions are skewed, the model may learn biased representations and perform poorly on the minority class.

In addition, due to patient privacy regulations (such as GDPR and HIPAA), sharing raw medical data between institutions is often not feasible~\cite{diagnostics13193140}. Federated Learning (FL) has been proposed as a promising solution to this problem. For instance, FL has been used to assist in cancer diagnosis~\cite{ma_assisted_2022} and several studies have reviewed its broader applications in medical imaging~\cite{guan2024federatedlearningmedicalimage, 1011453501813, doi:10.1200/CCI.20.00060}.

This section extends RHFL+ to the medical imaging domain to evaluate its performance in privacy-sensitive, real-world settings.

\subsection{Dataset Summary}

\hspace{1.5em}We consider three publicly available medical imaging datasets, summarized in Table~\ref{tab:dataset-summary}. These datasets were selected for their accessibility and relevance to breast cancer classification tasks. Table~\ref{tab:config_dataset} presents the corresponding training configurations used for each dataset, tuned for better performance given dataset characteristics.

For example, the CBIS-DDSM and BreastMNIST datasets are relatively small and thus prone to overfitting. Therefore, we use smaller batch sizes, learning rates and fewer training epochs. Additionally, since BreastMNIST is grayscale and resized to 32$\times$32, we chose MNIST as the public dataset for knowledge distillation in that experiment, rather than CIFAR-100, which uses 3-channel RGB images. Similarly, for 224$\times$224 three-channel CBIS-DDSM data after preprocessing, we use resized 224$\times$224 CIFAR-100. This ensures input format compatibility.

\begin{table}[H]
\centering
\resizebox{0.85\textwidth}{!}{%
\begin{tabular}{lccc}
\toprule
\textbf{Attribute} & \textbf{BreastMNIST~\cite{vabalas_machine_2019}} & \textbf{CBIS-DDSM~\cite{lee2016cbisddsm,lee_curated_2017}} & \textbf{BHI}~\cite{mooney2017breast} \\
\midrule
\textbf{Modality} & Ultrasound & Mammography & Histopathology \\
\textbf{Training Samples} & 546 & 607 & 10000(277524) \\
\textbf{Color Space} & Grayscale & Grayscale & RGB \\
\textbf{Original Size} & 64$\times$64 & Varies & 50$\times$50 \\
\textbf{Resized To} & 32$\times$32 & 224$\times$224 & 32$\times$32 \\
\textbf{Negative Label} & Normal/Benign & Benign & IDC Negative \\
\textbf{Positive Label} & Malignant & Malignant & IDC Positive \\
\bottomrule
\end{tabular}%
}
\caption{Summary of medical imaging datasets used in this study}
\label{tab:dataset-summary}
\end{table}

\vspace{2.5cm}

\begin{longtable}{@{}p{4.5cm}p{2.2cm}p{2.2cm}p{2.2cm}p{2.2cm}@{}}
\toprule
\textbf{Configuration} & \textbf{CBIS-DDSM (Benign/Malignant)} & \textbf{CBIS-DDSM (Density 1--4)} & \textbf{BreastMNIST} & \textbf{BHI} \\ 
\midrule
\textbf{Private Dataset Length} & 400 & 400 & 400 & 12500 \\
\textbf{Private Total Length} & 607 & 607 & 546 & 30000 \\
\textbf{Private Output Channels} & 2 & 4 & 2 & 2 \\
\textbf{Public Dataset Length} & 50 & 50 & 100 & 5000 \\
\textbf{Train Batch Size} & 8 & 8 & 8 & 256 \\
\textbf{Test Batch Size} & 8 & 8 & 8 & 512 \\
\textbf{Number of Rounds} & 20 & 20 & 20 & 40 \\
\textbf{Private Nets} & CNN224 $\times$ 4 & CNN224 $\times$ 4 & CNN $\times$ 4 & heterogeneous \\
\textbf{Public Dataset Name} & CIFAR-100 & CIFAR-100 & MNIST & CIFAR-100 \\
\textbf{Resized Image} & 224 & 224 & 32 & 32 \\
\textbf{Learning Rate} & 0.0001 & 0.0001 & 0.0001 & 0.001 \\
\textbf{Pretrain} & false & false & false & true/false \\
\textbf{Metric} & PR-AUC & ROC-AUC & PR-AUC & PR-AUC \\
\bottomrule
\caption{Configuration Summary of Medical Imaging Dataset Experiments}
\label{tab:config_dataset}
\end{longtable}

\subsection{Data Preprocessing}

\hspace{1.5em}In binary classification tasks, pairflip and symmetric label noise become functionally similar when the label space is limited. Therefore, we apply only pairflip noise during training to simulate real-world label corruption.

\subsubsection*{CBIS-DDSM (Raw)}

\hspace{1.5em}The CBIS-DDSM dataset contains two categories of mammograms: \textit{Calcification} and \textit{Mass Margin}. Additionally, each case includes both full-view and cropped images. For classification purposes, we focus on the craniocaudal (CC) view of the cropped images, as they are better localized and more relevant for classification tasks.

When converting raw DICOM files to training data, we take great care to prevent data leakage. Specifically, I ensure that images from the same patient do not appear in both the training and testing sets. Although the cropped images are typically used for segmentation due to the availability of lesion masks, I also utilize the masks during preprocessing. Any image whose mask has zero area is discarded, as it contains no useful lesion information for classification.

The dataset supports two distinct labeling schemes. The first is a binary classification task of benign and malignant cases. The second is a four-class classification task based on breast tissue density, with categories ranging from density level 1 to 4. 

\subsection{Model Architecture}

\hspace{1.5em}The original models used in RHFL+ (ResNet10, ResNet12 MobileNetV2, ShuffleNet) were optimized for datasets like CIFAR-10, which are relatively large and balanced. However, in our medical imaging tasks, datasets are significantly smaller and more imbalanced. Directly applying these complex models led to overfitting or unstable training. To mitigate these issues, we designed a shallow convolutional neural network (CNN) with fewer parameters and smaller receptive fields, tailored for small-scale datasets such as BreastMNIST. For higher-resolution datasets like CBIS-DDSM, which consist of 224$\times$224 images, we use an extended variant of the same architecture, referred to as Net224, which dynamically computes the flattened feature size to accommodate the larger input dimensions. Both models share a common backbone including five convolutional blocks, each of which integrates Batch Normalization, ReLU activation and MaxPooling. The head is a fully-connected multi-layer network, that output the target number of class predictions.

\subsection{Evaluation Metrics}

\hspace{1.5em}To assess model performance, we use more evaluation metrics than the one used in the original RHFL+ paper. Accuracy remains the primary metric. However, given the presence of class imbalance in many medical imaging datasets, we additionally use Precision-Recall Area Under the Curve (PR AUC) for binary classification tasks. This metric emphasizes on precision and recall, making it more informative when false negatives are particularly disfavored. For multi-class classification tasks, we use the Receiver Operating Characteristic Area Under the Curve (ROC AUC) to evaluate the model’s ability to distinguish between different classes. Together, these metrics offer a more balanced and nuanced evaluation of model performance.

\subsection{Sanity Check}

\hspace{1.5em}To validate the correctness of my training pipeline, I performed a sanity check before official training: training on a tiny balanced subset (10 samples per class). 

\begin{figure}[H]
    \centering
    \includegraphics[width=0.8\textwidth]{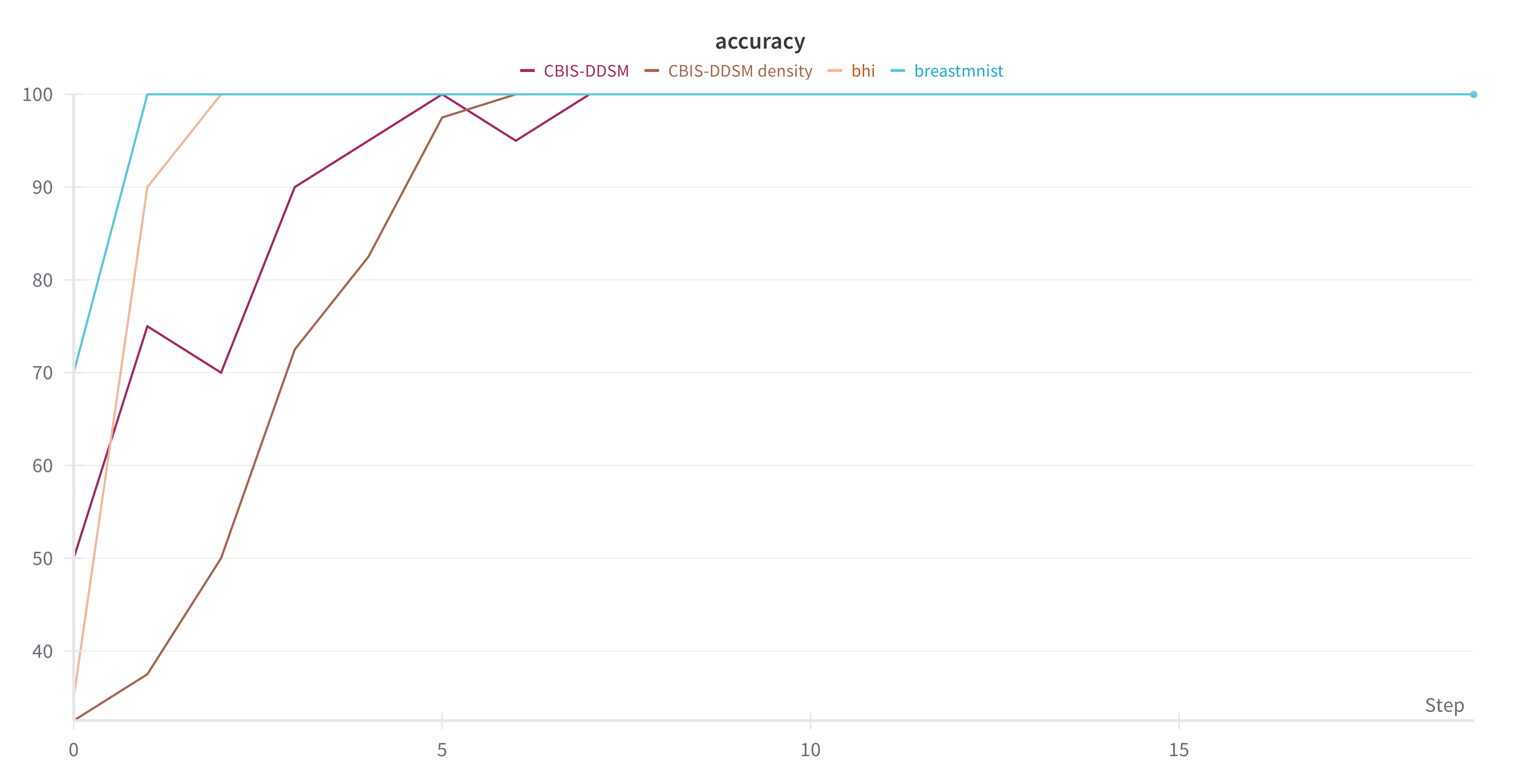}
    \caption{Sanity check accuracy of all medical imaging dataset, 20 iterations, 10 samples from each class, no noise}
    \label{fig:sanity_check}
\end{figure}

\begin{figure}[h]
    \centering
    \includegraphics[width=0.8\textwidth]{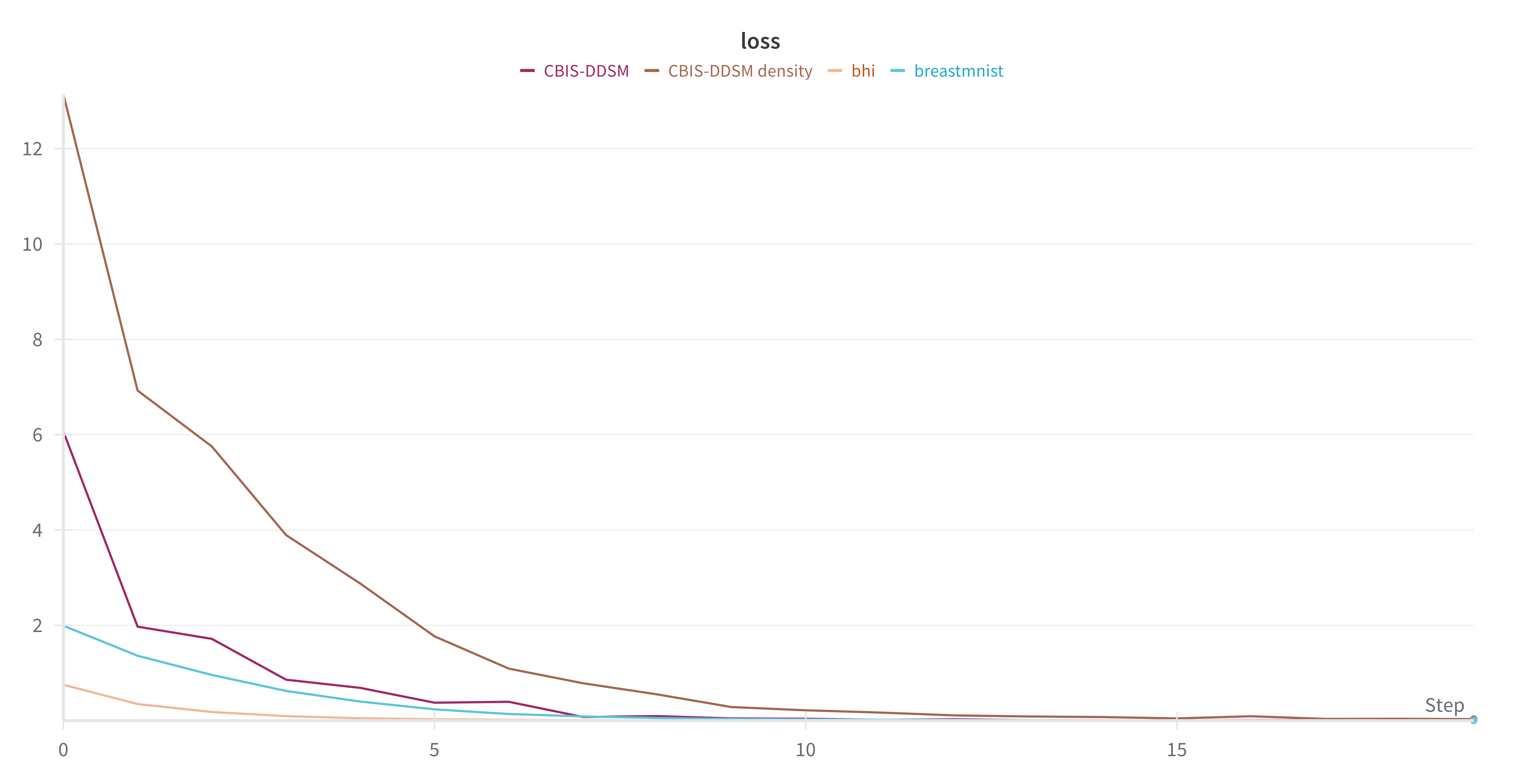}
    \caption{Sanity check loss of all medical imaging dataset, 20 iterations, 10 samples from each class, no noise}
    \label{fig:sanity_check_loss}
\end{figure}

According to the Figure\ref{fig:sanity_check} and Figure\ref{fig:sanity_check_loss}, we validate that all datasets reach near-perfect accuracy quickly (100\%), as expected. Loss decreases sharply for all datasets too. The sanity check validates the correctness of the data and pipeline implicitly.

\subsection{Experiment Setup}

\hspace{1.5em}All experiments use 40 communication rounds (or 20 for small datasets) with pairflip noise rates from 0.0 to 0.5. Training was conducted under identical hyperparameter settings unless otherwise stated and results are evaluated across the metrics above. All datasets have at least three algorithm applied: LocalOnly, CCR and ECCR(RHFL+). This is selected based on the reproduction results. Please refer to the Table\ref{tab:config_dataset} for more details of each dataset experiment setup.

While the original RHFL+ paper uses pretrained accuracy (after 40 rounds) as a baseline, we also consider a more meaningful baseline in our comparison: the \textbf{LocalOnly} strategy where applicable, where each client trains independently without collaboration. This helps to directly quantify the benefit of collaborative FL under the same number of communication rounds. It also helps when the dataset is too small to have pretrained models.

\subsection{Results}

\begin{figure}[h]
    \centering
    \includegraphics[width=\textwidth]{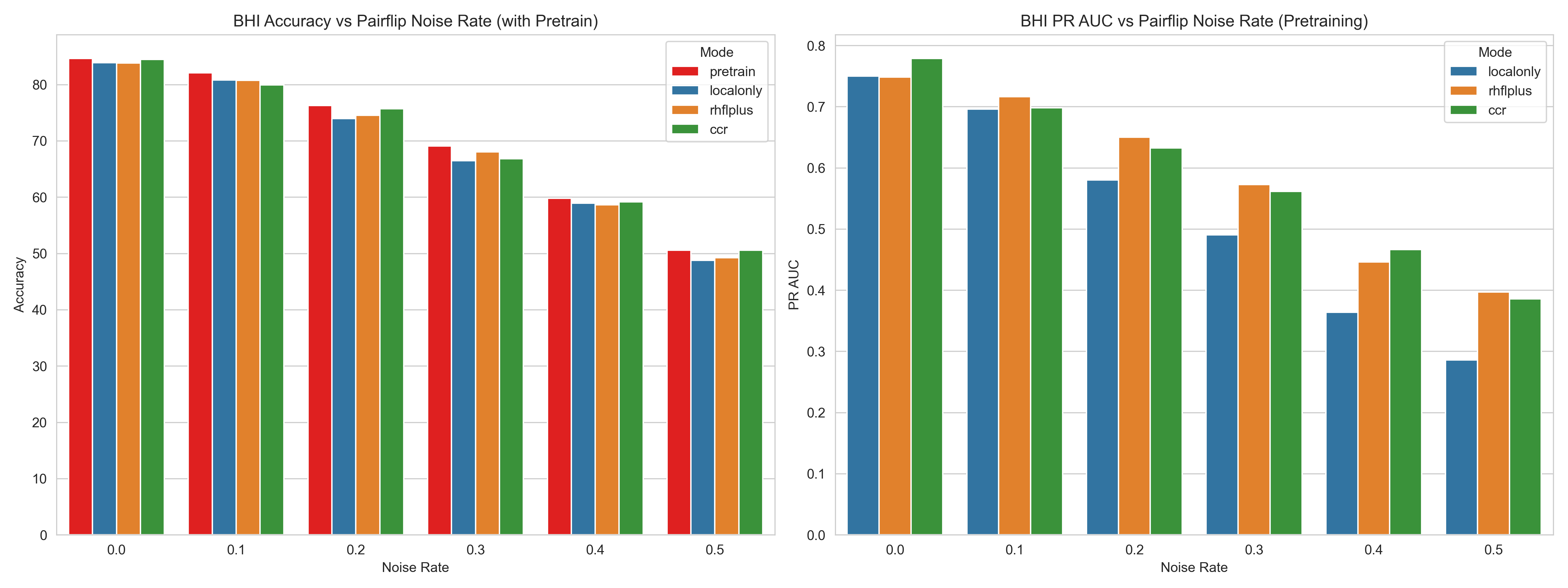}
    \caption{BHI as Private Dataset, with Pretrain, Pairflip noise rate $\mu = 0.0$ to $0.5$}
    \label{fig:pairflip_bhi}
\end{figure}

\begin{figure}[H]
    \centering
    \includegraphics[width=\textwidth]{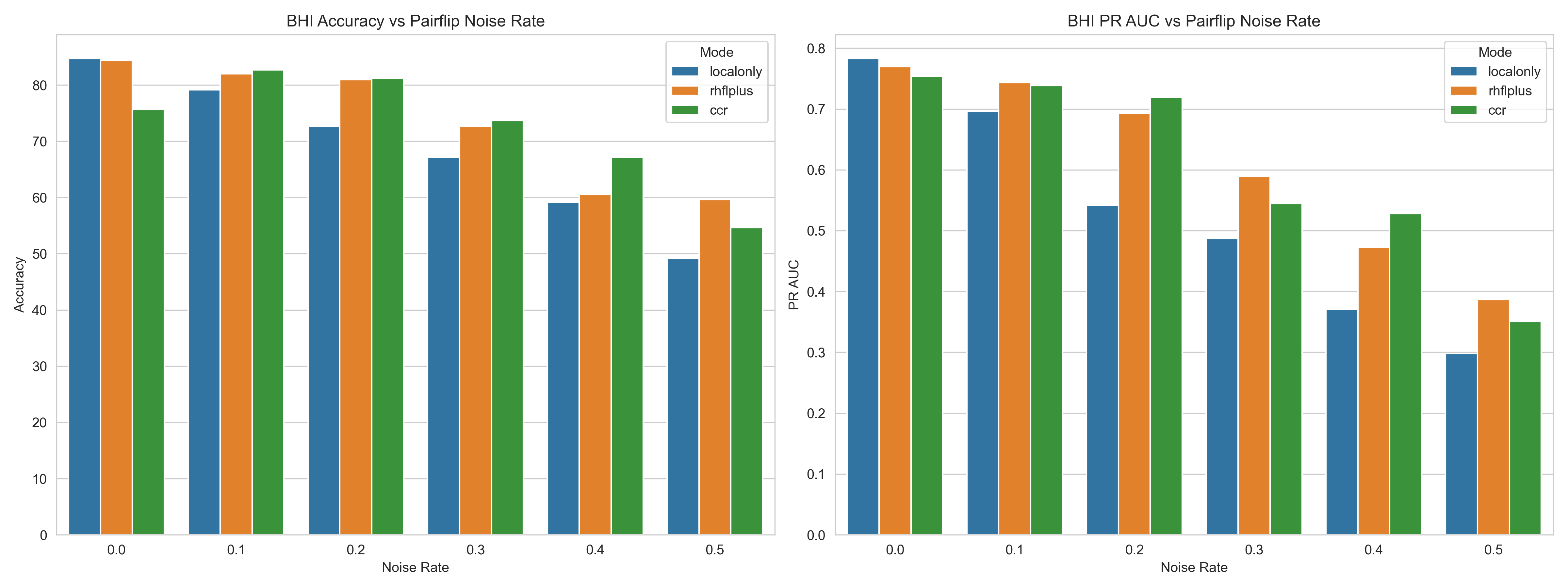}
    \caption{BHI as Private Dataset, no Pretrain, Pairflip noise rate $\mu = 0.0$ to $0.5$}
    \label{fig:pairflip_bhi_nopretrain_unbalanced}
\end{figure}

\paragraph{BHI}
Figures~\ref{fig:pairflip_bhi} and~\ref{fig:pairflip_bhi_nopretrain_unbalanced} show the average accuracy and PR AUC on the BHI dataset under increasing noise rates from $\mu = 0.0$ to $0.5$, comparing results with and without pretraining.

Figure~\ref{fig:pairflip_bhi} presents results using pretrained models. Interestingly and in contrast to the results on CIFAR-10, the LocalOnly baseline achieves higher accuracy than both federated learning methods. In Figure~\ref{fig:pairflip_bhi_nopretrain_unbalanced}, which shows results without pretraining, the RHFL-based methods perform better in most cases, suggesting that the pretrained models may be overfitting. This is further supported by the fact that the standalone pretrained model outperforms all other methods in Figure~\ref{fig:pairflip_bhi}. The PR AUC on the other hand, is observed to be better in RHFL-based method in most cases, which is consistent with and without pretraining.

As a result, we consider the no-pretrain setting a better reference point for comparing model performance due to reduced risk of overfitting. In Figure~\ref{fig:pairflip_bhi_nopretrain_unbalanced}, especially at higher noise rates, the RHFL-based methods consistently outperform distributed training without collaborative learning, with RHFL+ (with CCR) achieving the best performance in most scenarios. This highlights the potential benefits of RHFL-based approaches for medical imaging tasks under noisy conditions.

\begin{figure}[H]
    \centering
    \begin{minipage}{0.55\textwidth}
        \centering
        \includegraphics[width=\linewidth]{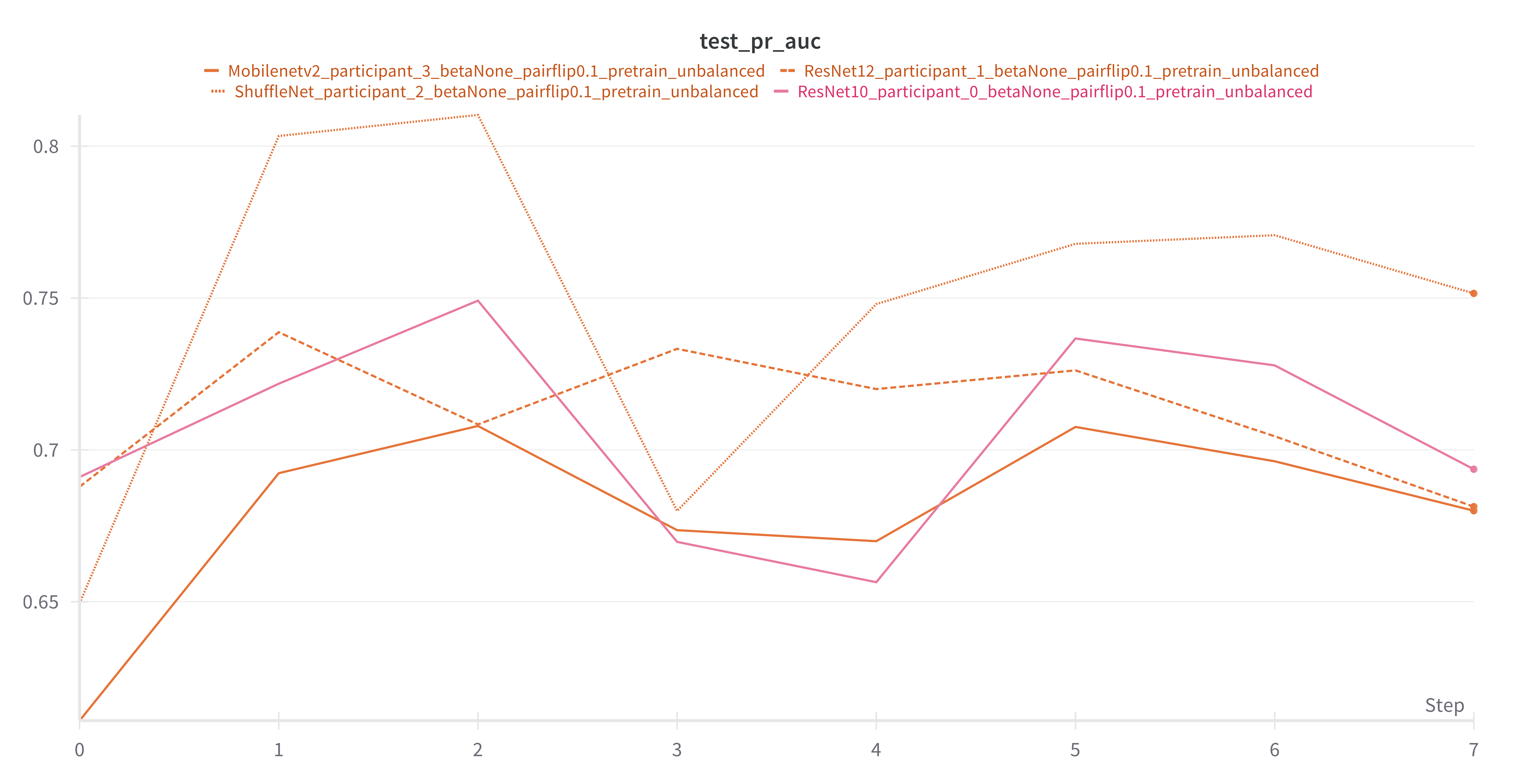}
        \vspace{1em}
        \includegraphics[width=\linewidth]{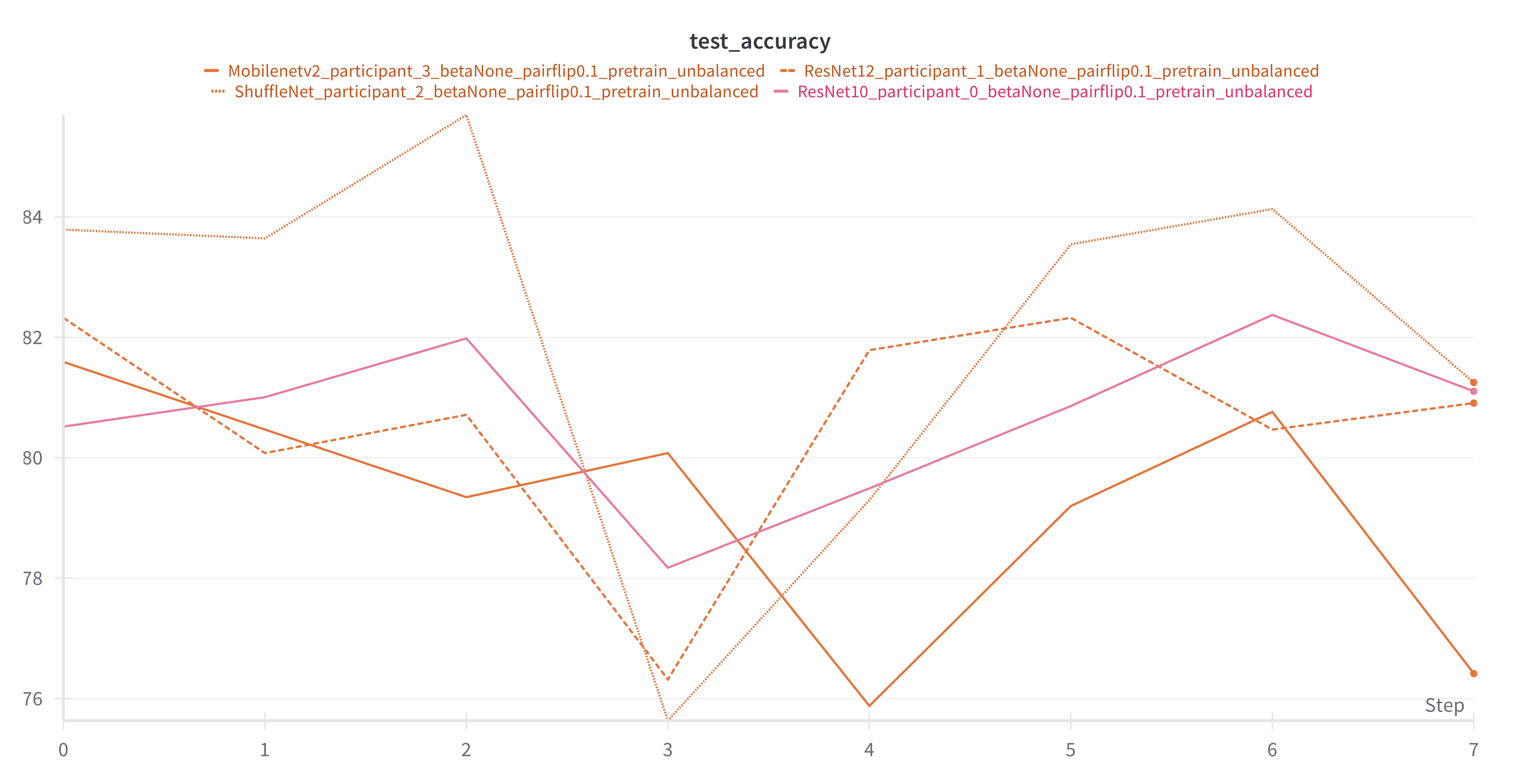}
    \end{minipage}
    \caption{PR AUC (top) and Test Accuracy(bottom), BHI as Private Dataset, with Pretrain, Pairflip noise rate 0.1}
    \label{fig:pretrain_bhi_details}
\end{figure}

\begin{figure}[H]
    \centering
    \begin{minipage}{0.55\textwidth}
        \centering
        \includegraphics[width=\linewidth]{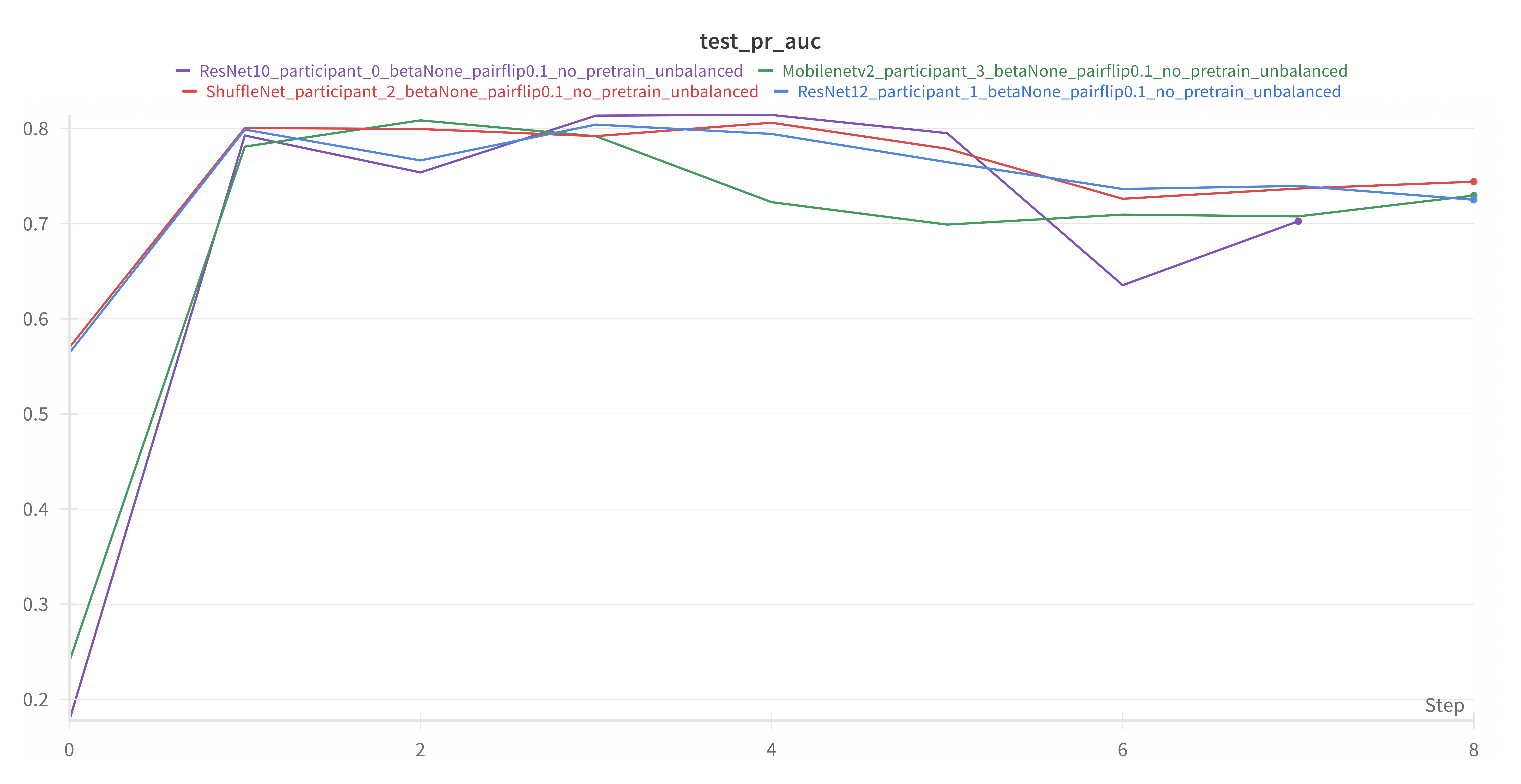}
        \vspace{1em}
        \includegraphics[width=\linewidth]{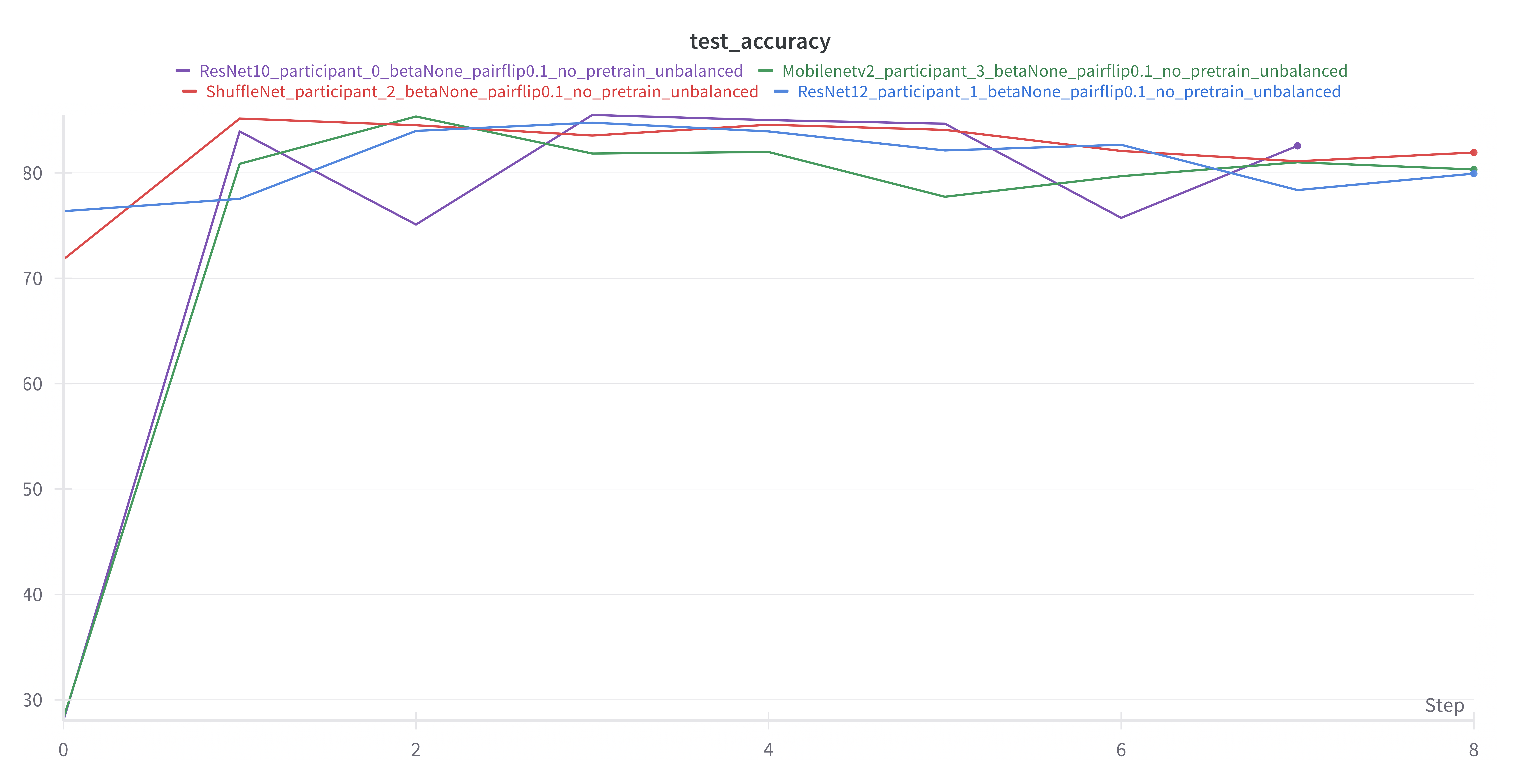}
    \end{minipage}
    \caption{PR AUC (top) and Test Accuracy(bottom), BHI as Private Dataset, no Pretrain, Pairflip noise rate 0.1}
    \label{fig:nopretrain_bhi_details}
\end{figure}

\begin{figure}[H]
    \centering
    \begin{minipage}{0.55\textwidth}
        \centering
        \includegraphics[width=\linewidth]{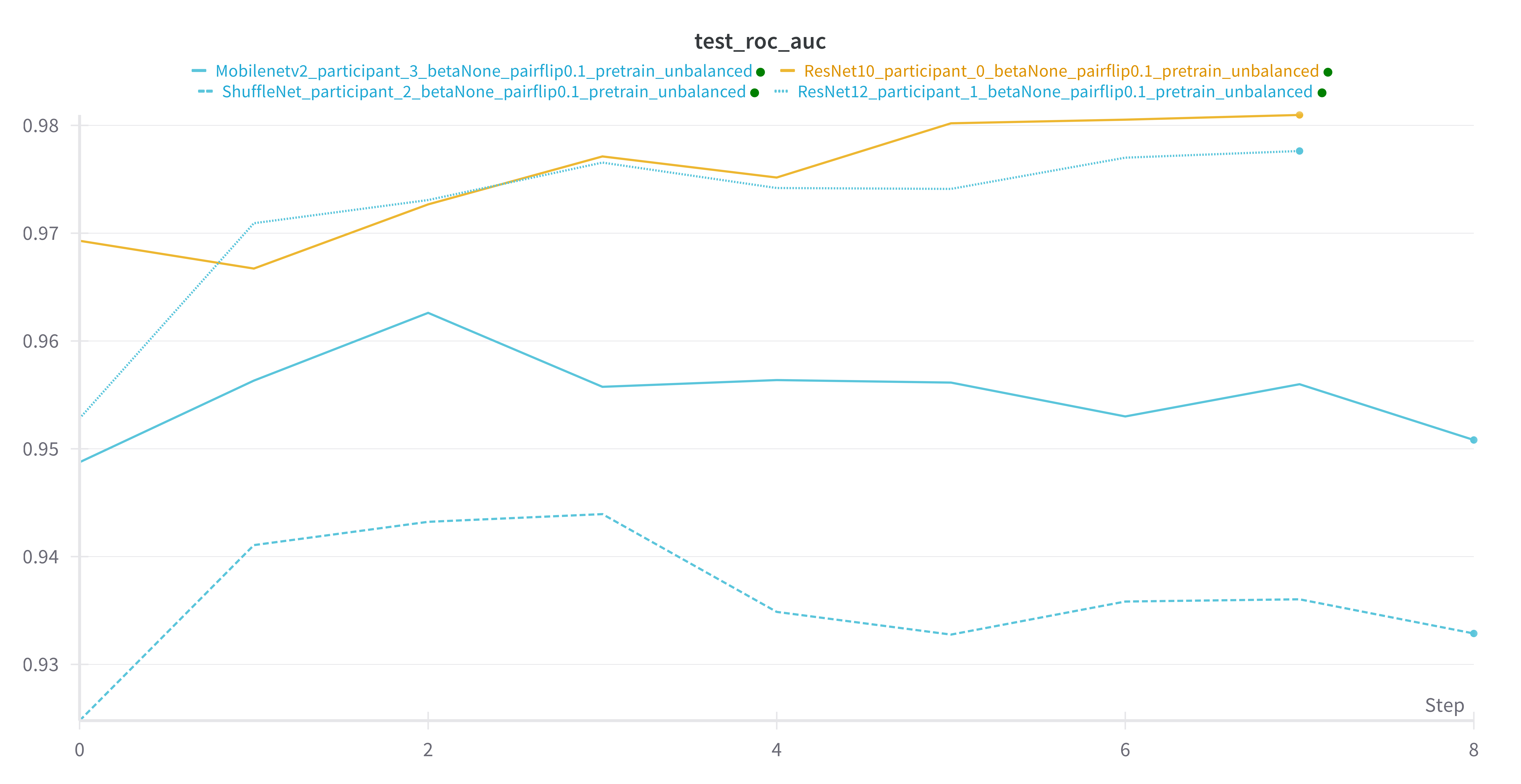}
        \vspace{1em}
        \includegraphics[width=\linewidth]{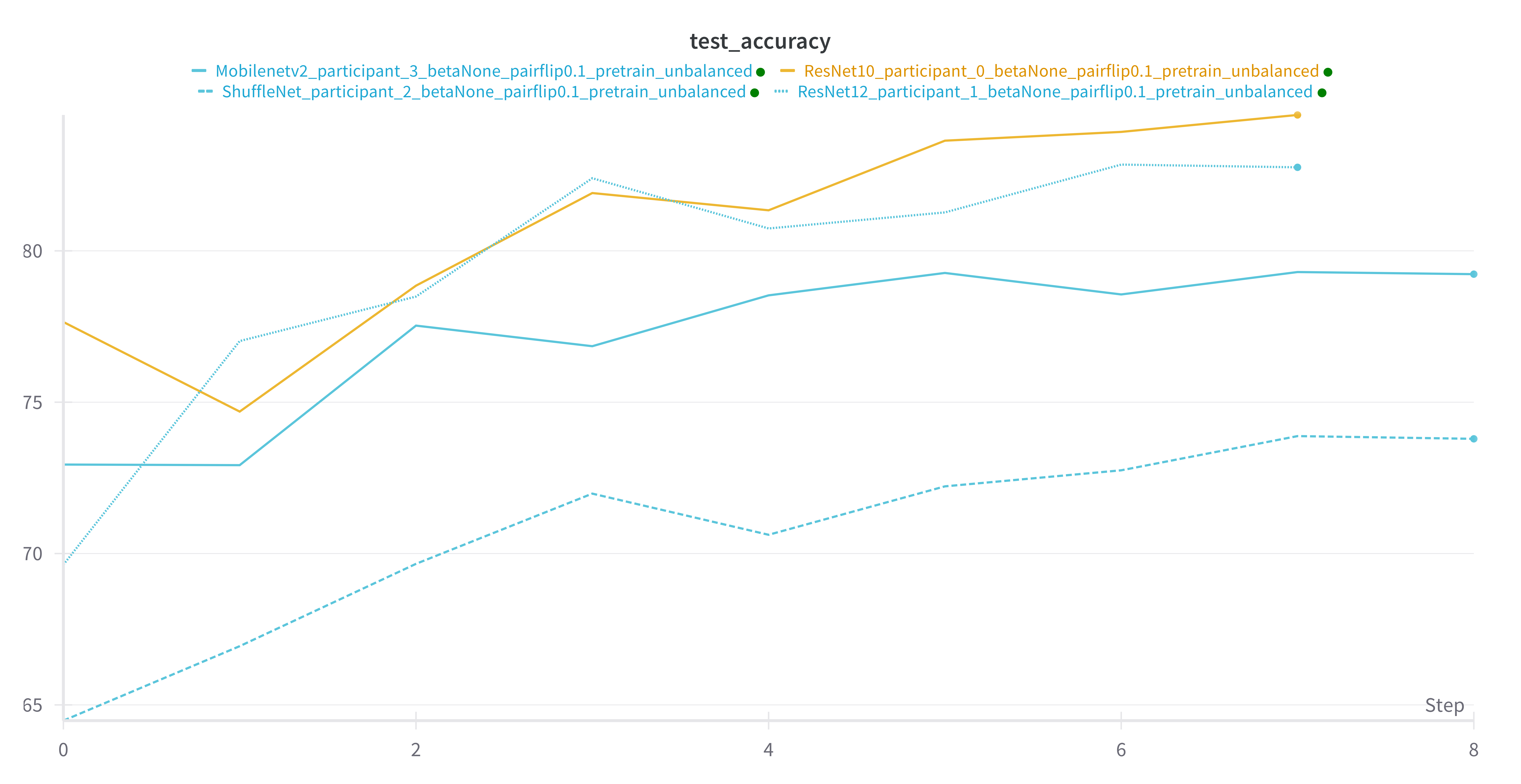}
    \end{minipage}
    \caption{PR AUC (top) and Test Accuracy(bottom), CIFAR-10 as Private Dataset, no Pretrain, Pairflip noise rate 0.1}
    \label{fig:nopretrain_cifar10_details}
\end{figure}

Figures~\ref{fig:pretrain_bhi_details} and~\ref{fig:nopretrain_bhi_details} further illustrate the trends in test accuracy and AUC every 5 epochs, with and without pretraining. These curves indicate that RHFL+ converges quickly without pretraining, but shows signs of overfitting and fluctuating when pretrained. This discrepancy might be dataset-dependent. For instance, as shown in Figure~\ref{fig:nopretrain_cifar10_details}, RHFL+ does not overfit on CIFAR-10 even with pretraining.

This suggests that RHFL-based methods may require careful hyperparameter tuning tailored to the specific dataset. In this project, the same parameter configuration from the original RHFL+ paper (optimized for CIFAR-10) was used on BHI due to time constraints, which might explain the observed overfitting in the pretrained case.

\begin{figure}[H]
    \centering
    \includegraphics[width=\textwidth]{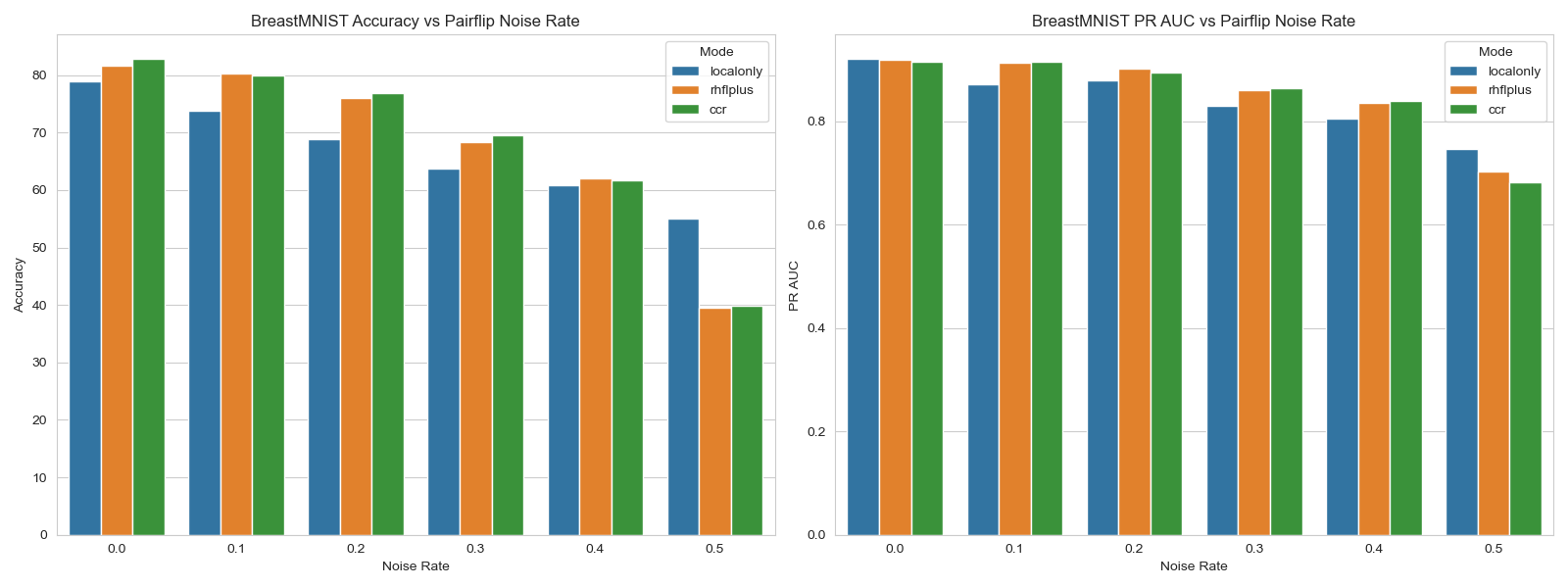}
    \caption{BreastMNIST as Private Dataset, no Pretrain, Pairflip noise rate $\mu = 0.0$ to $0.5$}
    \label{pairflip_breastmnist_nopretrain_unbalanced}
\end{figure}

\paragraph{BreastMNIST}
According to Figure~\ref{pairflip_breastmnist_nopretrain_unbalanced}, RHFL-based methods (RHFL+ and CCR) demonstrate robust performance on the BreastMNIST dataset under pairflip label noise. Across noise rates from 0.0 to 0.4, both methods consistently outperform the LocalOnly baseline. However, when the noise rate reaches 0.5, the performance of both RHFL-based methods drops sharply, falling below that of LocalOnly. This suggests that while the resilience of RHFL-based approaches degrades significantly under extremely high noise conditions in small dataset.

\paragraph{CBIS-DDSM Density and Binary}

\begin{figure}[H]
    \centering
    \includegraphics[width=\textwidth]{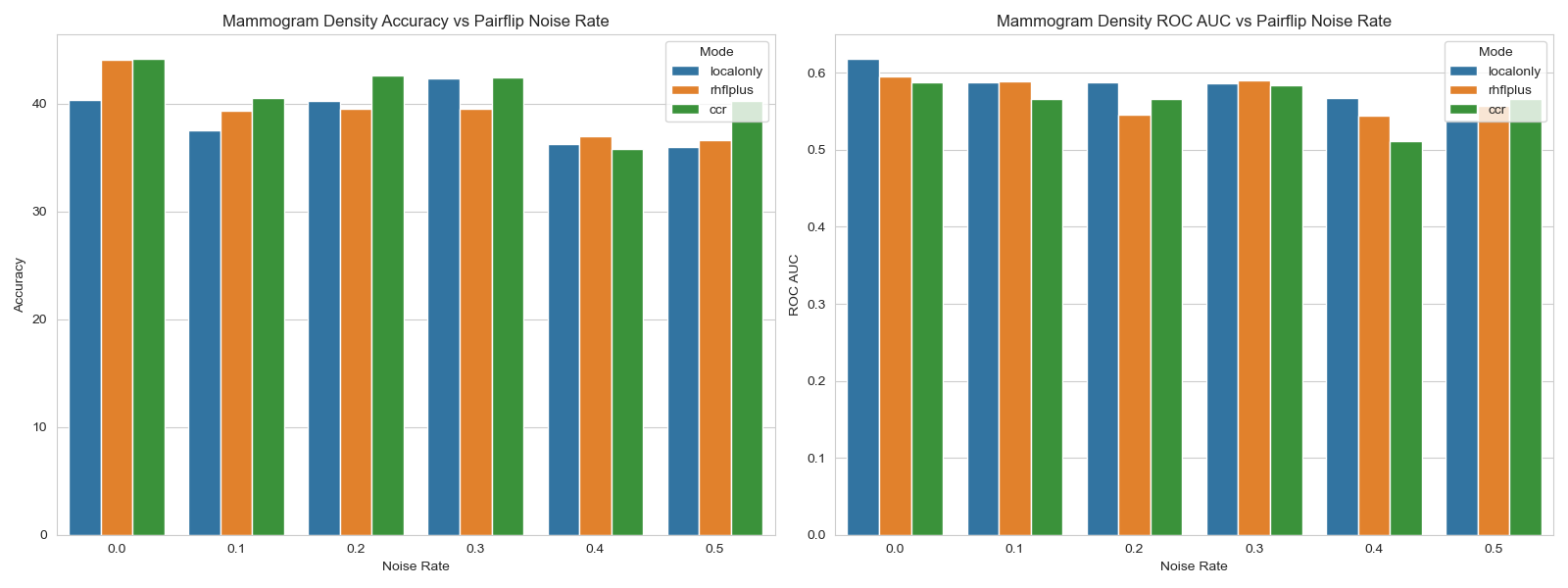}
    \caption{CBIS-DDSM as Private Dataset, Density as the label, with Pretrain, Pairflip noise rate $\mu = 0.0$ to $0.5$}
    \label{fig:mammogram_density_pairflip}
\end{figure}

\begin{figure}[H]
    \centering
    \includegraphics[width=\textwidth]{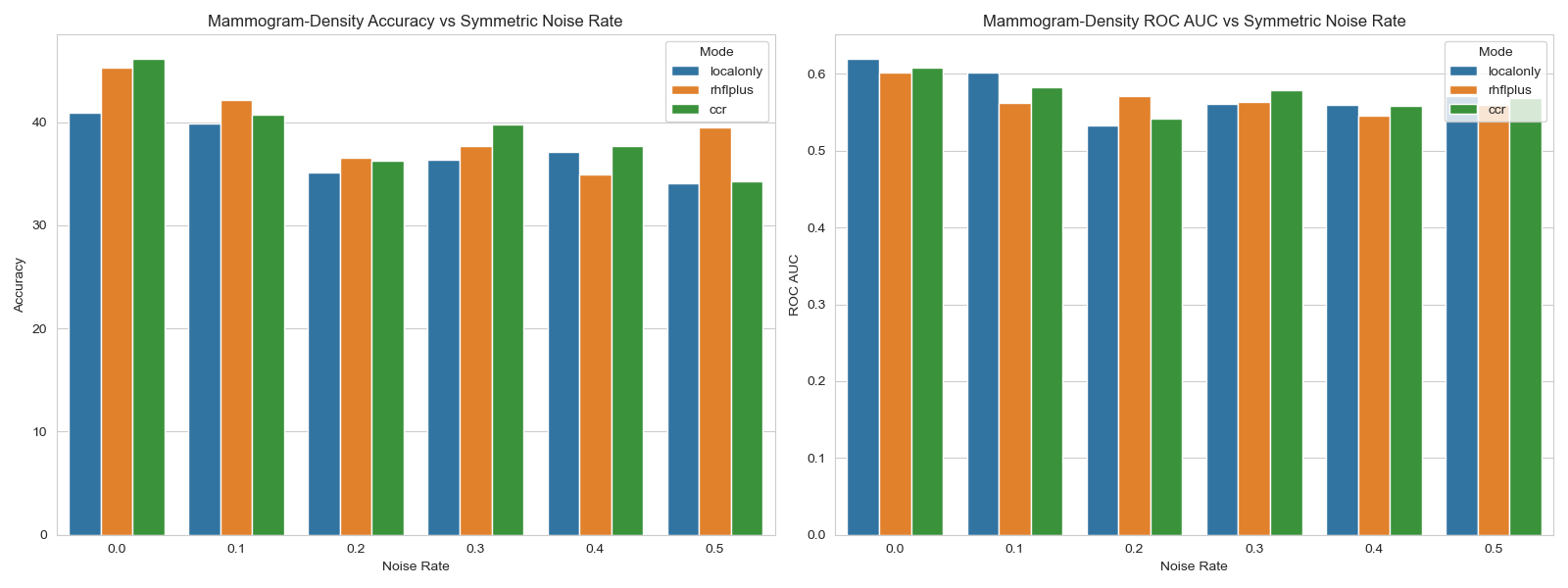}
    \caption{CBIS-DDSM as Private Dataset, Density as the label, with Pretrain, Symmetric noise rate $\mu = 0.0$ to $0.5$}
    \label{fig:mammogram_density_symmetric}
\end{figure}

\begin{figure}[h]
    \centering
    \includegraphics[width=\textwidth]{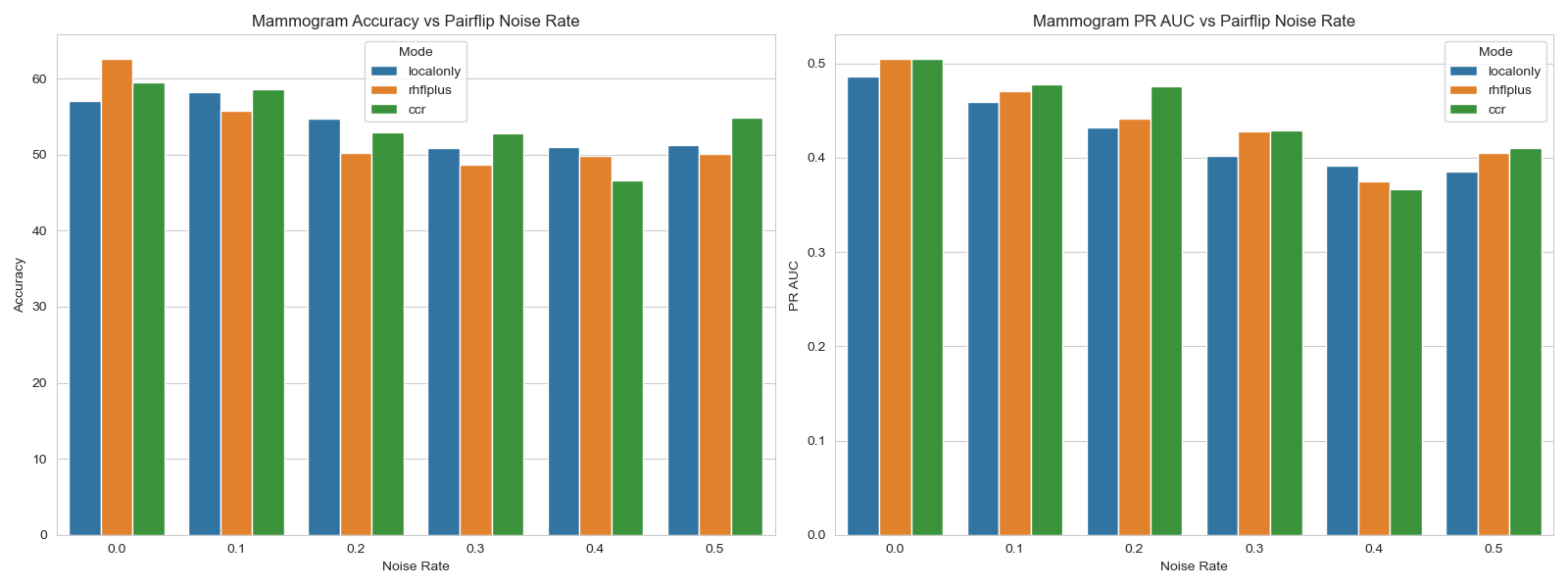}
    \caption{CBIS-DDSM as Private Dataset, Binary labels, no Pretrain, Pairflip noise rate $\mu = 0.0$ to $0.5$}
    \label{pairflip_mammogram_nopretrain_unbalanced}
\end{figure}

According to Figures~\ref{fig:mammogram_density_pairflip}, \ref{fig:mammogram_density_symmetric} and \ref{pairflip_mammogram_nopretrain_unbalanced}, the performance of RHFL+ appears inconsistent across different settings on the CBIS-DDSM dataset. Moreover, the overall classification accuracy is lower than expected. Since the dataset has passed the sanity checks, the underlying causes likely stem from characteristics intrinsic to the data. In general, the CBIS-DDSM dataset is not well-suited as a benchmark for evaluating RHFL+ under conditions of data imperfection.

\subsection{CBIS-DDSM New Model}
\begin{figure}[h]
    \centering
    \includegraphics[width=\textwidth]{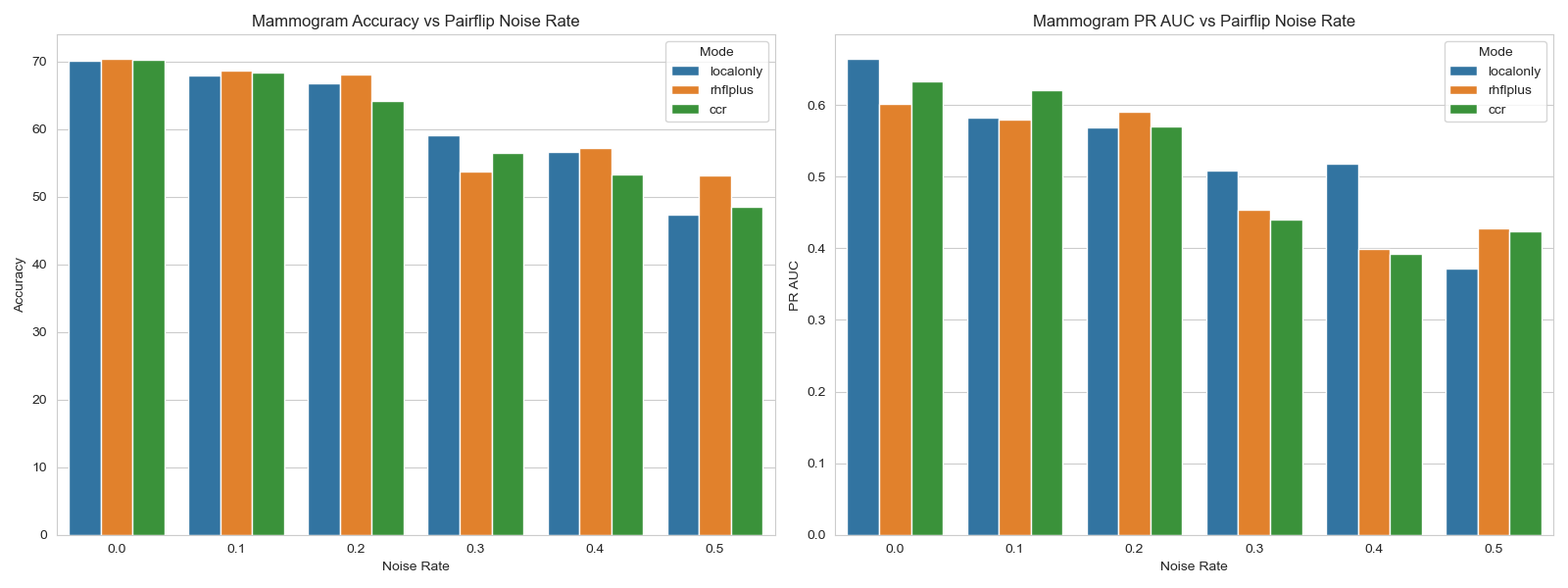}
    \caption{CBIS-DDSM as Private Dataset, New Model, Binary labels, no Pretrain, Pairflip noise rate $\mu = 0.0$ to $0.5$}
    \label{new_mammo}
\end{figure}

\hspace{1.5em}To improve performance on the CBIS-DDSM dataset, we adopted a more expressive model architecture by integrating a pretrained EfficientNet backbone with a lightweight embedding head and sigmoid classifier. The reason we use EfficientNet is due to its excellent transfer learning performance \cite{tan2020efficientnetrethinkingmodelscaling}. This design facilitates richer feature extraction, leveraging the pretrained backbone to address the challenge of the dataset being too small to train a deep network from scratch with sufficient generalization. The three-layer embedding head enables non-linear transformations tailored to the characteristics of the dataset. In addition, data augmentation using ColorJitter was applied to enhance the model’s robustness to variations in mammogram images. As shown in Figure~\ref{new_mammo}, the performance in terms of both test accuracy and PR AUC is noticeably improved compared to the previous shallow model results in Figure~\ref{pairflip_mammogram_nopretrain_unbalanced}. However, the advantages of RHFL+ were not fully realized with the default hyperparameters from the original paper. In some cases, the RHFL+ method performed only slightly better than the local-only baseline or even worse. This highlights the importance of careful hyperparameter tuning for effectively adapting RHFL+ to medical imaging tasks.

\section{Conclusion}
\label{sec:conclusion}
\hspace{1.5em}We reproduced and extended the RHFL+ framework under label noise. The implementation follows closely the original design, while also generalizing it to multiple real-world medical imaging datasets and experimental scenarios. The results show that RHFL+ consistently improves over benchmarks across most noise settings and datasets, particularly on CIFAR-10, BreastMNIST and BHI. However, performance on the CBIS-DDSM dataset is less consistent, possibly due to the dataset's intrinsic challenging imaging characteristics. Nevertheless, our experiments confirm the potential of RHFL+ as a robust FL method.

This work contributes to the topic by providing a fully modular and extensible RHFL+ implementation in the NVFlare framework, enabling reproducibility and experimentation across different model architectures, loss functions and aggregation strategies. In addition, through extensive ablation studies, comparison experiments and scaling experiments, we rigorously evaluated the effectiveness, robustness and scalability of RHFL+ across a wide range of conditions, including varying noise rates, model heterogeneity and dataset domains.

\section{Discussion and Future Work}
\label{sec:next}
\hspace{1.5em}While the current experimental framework provides meaningful insights into the performance of RHFL+ under noisy label conditions, several limitations and directions for future improvement remain.

First, the use of a centralized controller in RHFL+ introduces a single point of failure in the system. This architecture may compromise robustness and scalability in real-world deployments. Exploring the fault-tolerant mechanism of NVFlare could improve system resilience. Latest NVFlare tries to mitigate this by using a High Availability (HA) solution as described in the documentation\cite{nvflareHighAvailability}.

Additionally, hyperparameter tuning was not performed independently for each dataset. The same parameters from the original RHFL+ paper (tuned for CIFAR-10) were reused due to time constraints. Future work should involve dataset-specific tuning of key RHFL+ parameters to fully adapt the method to the unique characteristics of each medical imaging dataset.

To address the limited size and diversity of medical imaging datasets, we could also use advanced deep learning-based data augmentation techniques. A comprehensive benchmark evaluation for advanced data augmentation in medical imaging has been conducted recently\cite{qi2025mediaugexploringvisualaugmentation}. Mixup\cite{zhang2017mixup} is found to be a well-performed method. Such methods may alleviate overfitting and enhance generalization in small-scale medical imaging tasks.

Finally, all current experiments were conducted using NVFlare's simulation mode due to limited compute resources. Running RHFL+ in real production deployment mode with distributed clients will be a critical step to validate its practicality.

\appendix

\section*{Acknowledgments}
This work is based on research conducted for an MPhil thesis submitted to the University of Cambridge. I am grateful to my supervisor Dr. Joshua Kaggie (Department of Radiology, University of Cambridge) for his invaluable guidance and support throughout this research.
\section*{AI Assistance Disclosure}
During the preparation of this manuscript, the author used AI for assistance with LaTeX formatting, and proofreading.
\bibliographystyle{plain}
\bibliography{reference}

\end{document}